\documentclass[journal]{IEEEtran}
\usepackage{color}
\usepackage{float}
\usepackage{amssymb}
\usepackage{amsmath}
\usepackage{ragged2e}
\usepackage{subfiles}
\usepackage{multirow}
\usepackage{booktabs}
\usepackage{graphicx}
\usepackage{subfigure}
\usepackage{stfloats}
\usepackage[table]{xcolor}
\usepackage[switch]{lineno}
\usepackage[square, comma, sort&compress, numbers]{natbib}
\usepackage[colorlinks, linkcolor=red, anchorcolor=red, citecolor=green, urlcolor=blue]{hyperref}

\definecolor{mark}{rgb}{0,0,0}
\definecolor{new}{rgb}{0,0,0}

\usepackage{algorithm}
\usepackage{algpseudocode}

\usepackage[justification=justified]{caption}
\usepackage{enumitem}
\setlist[itemize]{leftmargin=2em}

\begin{document}

\title{Learning to Model Graph Structural Information on MLPs via Graph Structure Self-Contrasting}

\author{Lirong~Wu,~Haitao~Lin,~Guojiang~Zhao,~Cheng~Tan,~and~Stan Z. Li$^\dagger$,~\IEEEmembership{Fellow,~IEEE}%
\thanks{Lirong Wu, Haitao Lin, Guojiang Zhao, Cheng Tan, and Stan Z. Li are with the AI Lab, Research Center for Industries of the Future, Westlake University, Hangzhou 310024, China. E-mail: \{wulirong, linhaitao, zhaoguojiang, tancheng, stan.zq.li\}@westlake.edu.cn. $^\dagger$ Corresponding Author.}}

\maketitle

\begin{abstract}
Recent years have witnessed great success in handling graph-related tasks with Graph Neural Networks (GNNs). However, most existing GNNs are based on message passing to perform feature aggregation and transformation, where the structural information is \textbf{explicitly} involved in the forward propagation by coupling with node features through graph convolution at each layer. As a result, subtle feature noise or structure perturbation may cause severe error propagation, resulting in extremely poor robustness. In this paper, we rethink the roles played by graph structural information in graph data training and identify that message passing is not the only path to modeling structural information. Inspired by this, we propose a simple but effective \textit{\underline{G}raph \underline{S}tructure \underline{S}elf-\underline{C}ontrasting} (GSSC) framework that learns graph structural information without message passing. The proposed framework is based purely on Multi-Layer Perceptrons (MLPs), where the structural information is only \textbf{implicitly} incorporated as prior knowledge to guide the computation of supervision signals, substituting the explicit message propagation as in GNNs. Specifically, it first applies structural sparsification to remove potentially uninformative or noisy edges in the neighborhood, and then performs structural self-contrasting in the sparsified neighborhood to learn robust node representations. Finally, structural sparsification and self-contrasting are formulated as a bi-level optimization problem and solved in a unified framework. Extensive experiments have qualitatively and quantitatively demonstrated that the GSSC framework can produce truly encouraging performance with better generalization and robustness than other leading competitors. Codes are publicly available at: \url{https://github.com/LirongWu/GSSC.}
\end{abstract}


\IEEEpeerreviewmaketitle

\section{Introduction}
Recently, the emerging Graph Neural Networks (GNNs) have demonstrated their powerful capability in handling graph-related tasks \cite{liu2020towards,wu2020comprehensive,zhou2020graph,rossi2018deep,wu2024mape,wu2024learning}. Despite their great success, most existing GNNs are based on message passing, which consists of two key computations: (1) $\textrm{AGGREGATE}$: aggregating messages from its neighborhood, and (2) $\textrm{UPDATE}$: updating node representation from its representation in the previous layer and the aggregated messages. \textcolor{mark}{Due to the \emph{explicit coupling of node features and graph structural information} in the $\textrm{AGGREGATE}$ operation, subtle feature noise or structure perturbation may cause severe error propagation during message passing, resulting in extremely poor robustness \cite{zugner2018adversarial,zheng2020robust,feng2019graph,freitas2022graph}. There has been some pioneering work \citep{hu2021graph,luo2021distilling} delving into the necessity of message passing for modeling graph structural information. These methods adopt a pure MLP architecture that is free from feature-structure coupling, where structural information is only implicitly used to guide the computation of downstream supervision.} For example, Graph-MLP \citep{hu2021graph} designs a neighborhood contrastive loss to bridge the gap between GNNs and MLPs by implicitly utilizing the adjacency information. Besides, LinkDist \citep{luo2021distilling} directly distills self-knowledge from connected node pairs into MLPs without the need for aggregating messages. Despite their great progress, these MLP-based models still cannot match the state-of-the-art GNNs in terms of classification performance due to the underutilization of graph structural information. Therefore, \textit{``how to make better use of graph structural information without message passing"} is still a challenging problem.

The quality of graph structural information plays a very key crucial in various graph learning algorithms, making graph sparsification techniques, such as DropEdge \cite{rong2019dropedge}, STR-Sparse \cite{zheng2020robust} and GAUG \cite{zhao2020data}, etc., emerge as a common means to improve performance. The goal of graph sparsification is to filter out noisy structural information, i.e., finding an informative subgraph from the input graph by removing noisy edges. However, most existing graph sparsification methods are tailored for general GNNs, and \textcolor{mark}{they jointly optimize graph sparsification and GNN training by back-propagation of downstream supervision in an end-to-end manner. However, such an optimization may work for GNNs but is hard to be extended directly to MLP-based models}, where structural information has been used to guide the computation of downstream supervision, and it is not feasible to use downstream supervision in turn to optimize the structural sparsification. 

\begin{figure}[!tbp]
	\begin{center}
		\subfigure[Graph sparsification for GNN models]{\includegraphics[width=0.9\linewidth]{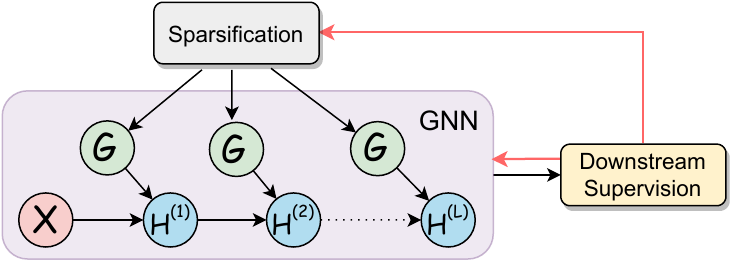}}
		\subfigure[Graph sparsification for MLP-based models (ours)]{\includegraphics[width=0.9\linewidth]{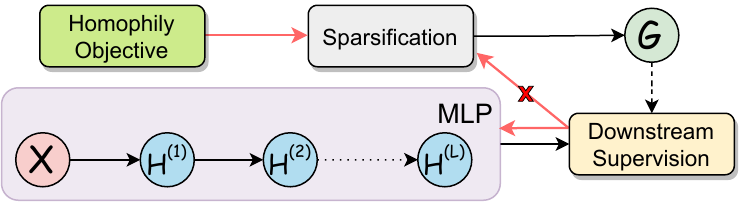}}
	\end{center}
        \vspace{-0.5em}
	\caption{A comparison between two cases of using graph sparsification for GNNs and MLP-based models. $\mathbf{X}$ denotes the input features, $\mathbf{H}$ denotes the hidden features, and $\mathbf{G}$ is the sparsified subgraph. The forward and backward propagation are marked as black and red lines, respectively.}
        \vspace{-0.5em}
	\label{fig:1}
\end{figure}

The differences between the two cases of using graph sparsification techniques for GNNs and MLP-based models are illustrated in Fig.~\ref{fig:1}, where the key point is \emph{whether graph structural information is used explicitly or implicitly}. \textcolor{mark}{For GNN models, the structural information $G$ is explicitly involved in the forward propagation, coupled with node features through graph convolution at each layer, and thus we can directly obtain the gradients, i.e., the derivative of supervision loss w.r.t the sparsification parameters. As a result, the parameters of GNN and graph sparsification can be jointly optimized by downstream supervision. In contrast, MLP-based models only implicitly utilize structural information $G$ to guide the computation of downstream supervision signals (denoted as a dashed line), and thus the derivative of supervision loss w.r.t the sparsification parameters is not available, which prevents the sparsification network from being directly optimized through downstream supervision (denoted as a red cross).} As an alternative, this paper proposes a homophily-oriented objective to guide the optimization of the sparsification network.

In this paper, we propose a simple yet effective \textit{\underline{G}raph \underline{S}tructural \underline{S}elf-\underline{C}ontrasting} (GSSC) framework, which consists of two main networks: \textit{(i) Structural Sparsification (STR-Sparse)} and \textit{(ii) Structural Self-Contrasting (STR-Contrast)}. In the proposed GSCC framework, we first apply the STR-Sparse network on the input graph to remove potentially uninformative or noisy edges in the neighborhood and then perform STR-Contrast in the sparsified neighborhood by imposing structural smoothness constraints between connected nodes. Specifically, the STR-Contrast network is based on MLPs, where structural information is only implicitly used to guide the computation of supervision signals but does not involve the forward propagation. Finally, we formulate structural sparsification and self-contrasting as a bi-level optimization problem and optimize the two networks using a tailor-made homophily-oriented objective and downstream supervision, respectively. Extensive experiments have shown that GSSC can produce truly encouraging performance with better generalization and robustness than other state-of-the-art competitors. To our best knowledge, we are the first work to explore the applicability of graph sparsification to (non-GNN) MLP-based models.

\section{Related Work}
\subsection{Graph Representation Learning} 
Recent years have witnessed the great success of GNNs in graph learning \cite{zhang2020deep,li2022ood,wu2023beyond,wu2022knowledge,wu2021graphmixup}. There are two categories of GNNs: spectral GNNs and spatial GNNs. The spectral-based GNNs define convolution kernels in the spectral domain based on the graph signal processing theory. For example, ChebyNet \cite{defferrard2016convolutional} uses the polynomial of the Laplacian matrix as the convolution kernel to perform message passing, and GCN is its first-order approximation. The spatial-based GNNs focus on the design of aggregation functions directly. For example, GraphSAGE \cite{hamilton2017inductive} employs a generalized induction framework to efficiently generate node embeddings for previously unseen data by aggregating known node features. GAT \cite{kipf2016semi}, on the other hand, adopts the attention mechanism to calculate the coefficients of neighbors for better information aggregation. We refer interested readers to the recent survey \cite{wu2020comprehensive} for more GNN variants, such as SGC \cite{wu2019simplifying}, APPNP \cite{klicpera2018predict} and DAGNN \cite{liu2020towards}. However, the above GNNs all share the de facto design that structural information is explicitly utilized for message passing to aggregate node features from the neighborhood.

Recently, there are some recent attempts to train graph data by combining contrastive learning and knowledge distillation \cite{,wu2023quantifying,wu2023extracting} with MLPs. For example, Graph-MLP \cite{hu2021graph} designs a neighborhood contrastive loss to bridge the gap between GNNs and MLPs by implicitly utilizing the adjacency information. Instead, LinkDist \cite{luo2021distilling} directly distills self-knowledge from connected node pairs into MLPs without the need to aggregate messages. Despite their great progress, they still cannot match the state-of-the-art GNN models in terms of classification performance, more importantly, they ignore the potential noise in the graph structure. \textcolor{new}{Another MLP-based model is Graph MLP-Mixer \cite{he2023generalization}, which generalizes ViT/MLP-Mixer to graph data and have achieved promising results.}

\textcolor{new}{Graph transformers (GTs) is another research area closely related to graph representation learning. For example, \cite{dwivedi2020generalization} is the first work that generalizes Transformer to graphs, which uses Laplacian position encoding to preserve local structural information. GraphiT \cite{mialon2021graphit} utilizes relative position encoding to enhance attention mechanism. Recently, GRIT \cite{ma2023graph} proposes a novel structural encoding called relative random walk probabilities (RRWP) to enhance local structure expressive power. Besides, Graph Transformers (SGFormer) \cite{lv2024sgformer} simplifies and empowers Transformers, which requires none of positional encodings, feature/graph pre-processing, or augmented loss. For more GT architectures, please refer to recent survey \cite{hoang2024survey}.}

\subsection{Graph Sparsification} 
The robust learning on graphs includes adversarial training \cite{feng2019graph,pan2019learning,jin2019latent}, label denoising \cite{xia2020towards}, structure learning \cite{jiang2019semi,klicpera2019diffusion,yu2020graph}, etc., among which the closest one to ours is graph sparsification, which aims to find a small subgraph from the input graph that best preserve some desired properties. Existing graph sparsification techniques can be divided into two categories: unsupervised and supervised. The unsupervised sparsification techniques, such as Spectral Sparsifier (SS) \cite{sadhanala2016graph} and Rank Degree (RD) \cite{voudigari2016rank}, mainly deal with simple graphs by some pre-defined graph metrics, e.g., node degree distribution, clustering coefficient, etc. Besides, DropEdge \cite{rong2019dropedge} randomly removes a fraction of edges according to the pre-defined probability before each training epoch to get sparsified graphs. In contrast, supervised sparsification directly parameterizes the sparsification process and optimizes it end-to-end along with GNN parameters under downstream supervision. For example, Learning Discrete Structure (LDS) \cite{franceschi2019learning} works under a transductive setting and learns Bernoulli variables associated with individual edges. Besides, GAUG \cite{zhao2020data} first optimizes the graph structure learning and GNN parameters in an end-to-end manner and then directly removes some low-importance edges during training. Moreover, NeuralSparse \cite{zheng2020robust} proposes a general framework that simultaneously learns to remove task-irrelevant edges and node representations by downstream supervision. \textcolor{new}{Furthermore, L2A \cite{wu2023learning} unifies graph sparsification (augmentation) and GNN training in a variational inference framework, which is applicable to both homophily and heterophily graphs.} However, these supervised sparsification techniques may be hard to be directly extended to existing MLP-based models.

\textcolor{mark}{\subsection{Graph Structure Learning}}
\textcolor{mark}{Recent advances in Graph Structure Learning (GSL) provide new insights into reducing the dependency on the given graph structure. The primary goal of graph structure learning is to infer an underlying graph structure from node features and then apply a GNN classifier to the inferred graph \cite{wu2023homophily}. For example, \cite{wu2023homophily} proposes \textit{Homophily-Enhanced Self-supervision for Graph Structure Learning} (HES-GSL), a method that provides additional self-supervision for learning an underlying graph structure. Similarly, \cite{gu2023homophily} proposes a novel method called Homophily-enhanced structure Learning for graph clustering (HoLe), based on the observation that subtly enhancing the degree of homophily within the graph structure can significantly improve GNNs and clustering outcomes. To address the under-supervision problem, \textit{Simultaneous Learning of Adjacency and GNN Parameters with Self-supervision} (SLAPS) \cite{fatemi2021slaps} proposes a feature reconstruction-based pretext task to provide more self-supervision for graph structure learning. Besides, RDGSL \cite{zhang2023rdgsl} proposes dynamic graph structure learning, where dynamic graph filters are designed to address the noise dynamics issue. Moreover, CGI \cite{wei2022contrastive} proposes
a contrastive graph structure learning via information bottleneck for
recommendation, which adaptively learns whether to drop an edge or node to obtain optimized structures in an end-to-end manner. We refer interested readers to a recent survey \cite{zhu2021deep} for more methods.}

\vspace{1em}
\section{Methodology}
\textbf{Used Notations.}
Given a graph $\mathcal{G}=(\mathcal{V}, \mathcal{E})$, where $\mathcal{V}$ is the set of $N$ nodes with features $\mathbf{X}=\left[\mathbf{x}_{1}, \mathbf{x}_{2}, \cdots, \mathbf{x}_{N}\right]\in \mathbb{R}^{N \times d}$ and $\mathcal{E}$ denotes the edge set. Each node $v_i \in \mathcal{V}$ is associated with a $d$-dimensional features vector $\mathbf{x}_{i}$, and each edge $e_{i, j} \in \mathcal{E}$ denotes a connection between node $v_i$ and $v_j$. The graph structure can also be denoted by an adjacency matrix $\mathbf{A} \in[0,1]^{N \times N}$ with $\mathbf{A}_{i,j}=1$ if $e_{i,j}\in\mathcal{E}$ and $\mathbf{A}_{i,j}=0$ if $e_{i,j} \notin \mathcal{E}$. Node classification is a typical node-level task where only a subset of node $\mathcal{V}_L$ with corresponding labels $\mathcal{Y}_L$ are known, and we denote the labeled set as $\mathcal{D}_L=(\mathcal{V}_L,\mathcal{Y}_L)$ and unlabeled set as $\mathcal{D}_U=(\mathcal{V}_U,\mathcal{Y}_U)$, where $\mathcal{V}_U=\mathcal{V} \backslash \mathcal{V}_L$. The task of node classification aims to learn a mapping $p(Y \mid \mathbf{X},\mathbf{A})$ on labeled data $\mathcal{D}_L$, so that it can be used to infer the label $Y \in \mathcal{Y}_U$.

\begin{figure*}[!htbp]
	\begin{center}
		\includegraphics[width=0.83\linewidth]{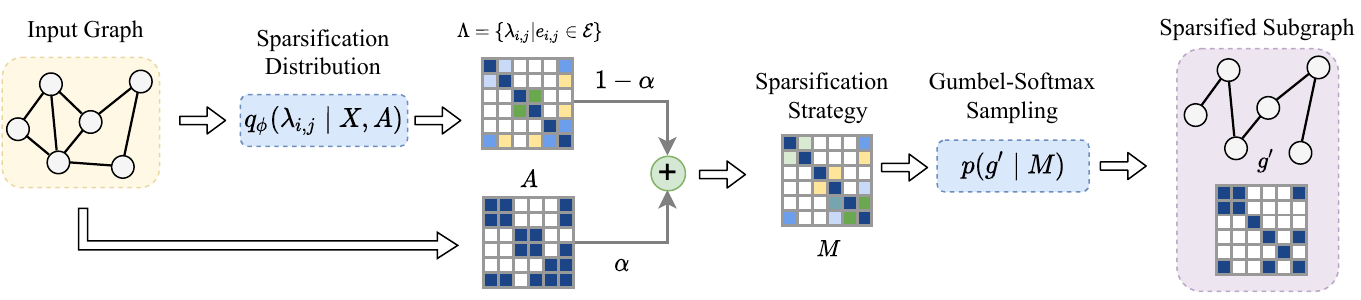}
	\end{center}
	\caption{Illustration of the proposed structural sparsification network, which consists of three main components: (1) Estimate $\Lambda = \{\lambda_{i,j} | i\in\mathcal{V},j\in\mathcal{N}_i \}$ by sparsification distribution $q_{\phi}\left(\lambda_{i,j} \mid \mathbf{X}, \mathbf{A}\right)$; (2) Obtain the sparsification strategy $\mathbf{M}$ by the weighted fusion; (3) Sample a sparsified subgraph $g^\prime$ from sparsification strategy $\mathbf{M}$ through Gumbel-Softmax sampling.}
	\label{fig:2}
\end{figure*}

\subsection{Theoretical Justification} \label{sec:3.1}
From the perspective of statistical learning, the key of node classification is to learn a mapping $p(Y \mid \mathbf{X},\mathbf{A})$ based on node features $\mathbf{X}$ and adjacency matrix $\mathbf{A}$. However, instead of directly working with the original graph, we would like to leverage sparsified subgraphs to remove task-irrelevant information and learn more robust representations. In other words, we are interested in the variant as follows
\begin{equation}
p(Y \mid \mathbf{X},\mathbf{A}) = \sum_{g\in\mathbb{S}_{\mathcal{G}}} p(Y \mid \mathbf{X}, g) p(g \mid \mathbf{X},\mathbf{A}),
\end{equation}
where $g\in\mathbb{S}_{\mathcal{G}}$ is a sparsified subgraph of original graph $\mathcal{G}$. In practice, the distribution space size of $\mathbb{S}_{\mathcal{G}}$ is $2^{|\mathcal{E}|}$, and it is intractable to enumerate all possible $g$ as well as estimate the exact values of $p(Y \mid \mathbf{X}, g)$ and $p(g \mid \mathbf{X},\mathbf{A})$. Therefore, we turn to approximate the distributions by two tractable parameterized functions $q_\theta(\cdot)$ and $q_\phi(\cdot)$, as follows
\begin{equation}
p(Y \mid \mathbf{X},\mathbf{A}) = \sum_{g\in\mathbb{S}_{\mathcal{G}}} q_\theta(Y \mid \mathbf{X}, g) q_\phi(g \mid \mathbf{X},\mathbf{A}),
\end{equation}
where $q_\theta(\cdot)$ and $q_\phi(\cdot)$ are approximation functions for $p(Y \mid \mathbf{X}, g)$ and $p(g \mid \mathbf{X},\mathbf{A})$. Moreover, to make the above graph sparsification process differentiable, we employ the commonly used reparameterization tricks \cite{jang2016categorical} to transform the discrete combinatorial optimization problem to a continuous probabilistic generative model, as follows
\begin{equation}
\sum_{g \in \mathbb{S}_{\mathcal{G}}} q_\theta(Y \mid \mathbf{X}, g) q_\phi(g \mid \mathbf{X},\mathbf{A}) \propto \sum_{g^{\prime} \sim q_\phi(g \mid \mathbf{X},\mathbf{A})} q_\theta(Y \mid \mathbf{X}, g^{\prime}),
\label{equ:3}
\end{equation}
where $g^{\prime}$ is a subgraph sampled from $q_\phi(g \mid \mathbf{X},\mathbf{A})$. 

To optimize Eq.~(\ref{equ:3}), a bi-level optimization framework is adopted to alternate between learning $q_\phi(g \mid \mathbf{X},\mathbf{A})$ and $q_\theta(Y \mid \mathbf{X}, g^\prime)$. In addition to the downstream supervision for learning $q_\theta(Y \mid \mathbf{X}, g^\prime)$, another objective function $\mathcal{H}(\cdot)$ is required to optimize $q_\phi(g \mid \mathbf{X},\mathbf{A})$. Intuitively, $\mathcal{H}(\cdot)$ should be an unsupervised metric to evaluate the quality of the sparsified graph $g^{\prime}$. Graph homophily, as an important graph property, may be a desirable option for $\mathcal{H}(\cdot)$.
\newline

\vspace{-0.5em}
\noindent \textbf{\textit{Introduction on Graph Homophily:}} The graph homophily is defined as the fraction of inter-class edges in a graph,
\begin{equation}
\begin{small}
    r=\frac{\left|\left\{(i, j):e_{i,j} \in \mathcal{E} \wedge y_{i}=y_{j}\right\}\right|}{|\mathcal{E}|}.
\end{small}
\end{equation}

\noindent The homophily ratio of the sparsified subgraph $g^{\prime}$ may be a desirable option for $\mathcal{H}(\cdot)$ for the following three reasons: \textit{(1)} Most common graphs adhere to the principle of homophily, i.e., ``birds of a feather flock together", which suggests that connected nodes often belong to the same class, e.g., friends in social networks often share the same interests or hobbies. \textit{(2)} It has been shown in previous work \cite{zhu2020beyond} that common GNN classifiers, such as GCN and GAT, usually perform better on datasets with higher homophily ratios. \textit{(3)} When downstream supervision is not accessible, the prior knowledge of graph homophily can serve as strong guidance for searching the suitable sparsified subgraph.
\newline

\vspace{-0.5em}
\noindent \textbf{\textit{Problem Statement:}} In this paper, the three important issues on framework design can be summarized as:

\begin{itemize}
    \item Implementing sparsification network $q_\phi(g \mid \mathbf{X},\mathbf{A})$ that takes node features $\mathbf{X}$ and adjacency matrix $\mathbf{A}$ as inputs to generate a sparsified subgraph $g^\prime$. (Sec.~\ref{sec:3.2})
    \item Implementing self-contrasting network $q_\theta(Y \mid \mathbf{X}, g^\prime)$ that takes node features $\mathbf{X}$ and the sparsified subgraph $g^\prime$ as inputs to make predictions $Y$. (Sec.~\ref{sec:3.3})
    \item Formulating the sparsification and self-contrasting in a bi-level optimization framework and proposing a homophily-oriented objective function (Sec.~\ref{sec:3.4})
\end{itemize}

\subsection{Structural Sparsification Network} \label{sec:3.2}
The sparsification network aims to generate a discrete sparsified subgraph $g^\prime$ for graph $\mathcal{G}$ as shown in Fig.~\ref{fig:2}, serving as the approximation function $q_\phi(g \mid \mathbf{X},\mathbf{A})$. Therefore, we need to answer three questions about the sparsification network: \textit{(1)} How to model the sparsification distribution? \textit{(2)} How to \textbf{differentiably} sample discrete sparsified subgraphs from the learned sparsification distribution? \textit{(3)} How to optimize the sparsification distribution and subgraph sampling process? Next, we first answer \textit{Question (1)(2)} and defer the discussion of optimization strategy until Sec.~\ref{sec:3.4}.

\subsubsection{\textbf{Sparsification Distribution}} To model the sparsification distribution, we introduce a set of latent variables $\Lambda = \{\lambda_{i,j} | e_{i,j}\in\mathcal{E}\}$, where $\lambda_{i,j}\in[0,1]$ denotes the sparsification probability between node $v_i$ and $v_j$. We can estimate $\lambda_{i,j}$ directly with distribution $q_{\phi}\left(\lambda_{i,j} \mid \mu_{i,j}\right)$ prameterized by $\mu_{i,j} \in \mathbb{R}^F$. However, directly fitting each $q_{\phi}\left(\lambda_{i,j} \mid \mu_{i,j}\right)$ locally involves solving $|\mathcal{E}|F$ parameters, which increases the over-fitting risk given the limited labels in the graph. Thus, we consider the amortization inference \cite{kingma2013auto}, which avoids the optimization of parameter $\mu_{i,j}$ for each local probability distribution $q_{\phi}\left(\lambda_{i,j} \mid \mu_{i,j}\right)$ and instead fits a shared neural network to model parameterized posterior. Specifically, we first transform the input to a low-dimensional hidden space, done by multiplying the features of input nodes with a shared parameter matrix $\mathbf{W} \in \mathbb{R}^{F \times d}$, that is, $\mathbf{z}_i=\mathbf{W} x_i$. Then, we parameterize the sparsification distribution $\lambda_{i,j}$ as follows:
\begin{equation}
q_{\phi}\left(\lambda_{i,j} \mid \mathbf{X}, \mathbf{A}\right)=\sigma\left(\mathbf{z}_i \mathbf{z}_j^{T}\right),
\label{equ:5}
\end{equation}
where $\sigma(\cdot)$ is an element-wise sigmoid function.

\subsubsection{\textbf{Subgraph Sampling}} To sample discrete sparsified subgraphs from the learned sparsification distribution and make the sampling process differentiable, we adopt Gumbel-Softmax sampling \cite{jang2016categorical}, which can be formulated as follows
\begin{equation}
\hspace{-1em}
g^{\prime}_{i,j}=\Bigg\lfloor\frac{1}{1+\exp^{-\big({\log \mathbf{M}_{i,j}+G}\big)/\tau}}+\frac{1}{2}\Bigg\rfloor, \quad 
 \text{where} \ i\in\mathcal{V},j\in\mathcal{N}_i
\label{equ:6}
\end{equation}
where $\mathbf{M}_{i,j}=(1-\alpha) q_{\phi}\left(\lambda_{i,j} \mid \mathbf{X}, \mathbf{A}\right)+\alpha\mathbf{A}_{i,j}$ is defined as the learned \emph{structural sparsification strategy}. In addition, $\alpha\in[0,1]$ is the fusion factor, which aims to prevent the sampled sparsified subgraph $g^\prime$ from deviating too much from the original graph. Besides, $\tau$ is the distribution temperature, and $G \sim \text{Gumbel}(0,1)$ is a gumbel random variate. 

Next, we will discuss in detail how to model the STR-Contrast network $q_\theta(Y \mid \mathbf{X}, g^\prime)$ based on the node features $\mathbf{X}$ and the sampled sparsified subgraph $g^\prime$. Without loss of generality, we can denote the edge set of the sparsified graph $g^\prime$ as $\mathcal{E}^\prime_g$ to distinguish $\mathcal{E}$ of the original graph $\mathcal{G}$. 

\begin{figure*}[!htbp]
    \vspace{-2em}
	\begin{center}
		\includegraphics[width=0.85\linewidth]{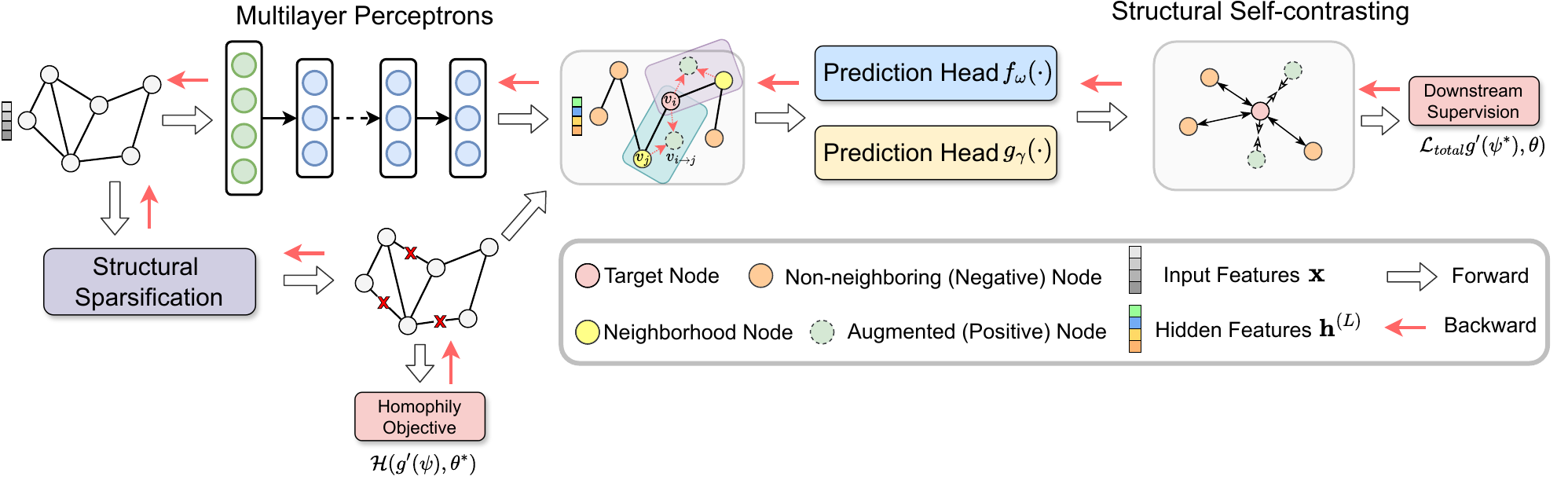}
	\end{center}
    \vspace{-0.5em}
	\caption{Illustration of the proposed GSSC framework, consisting of a structural sparsification network, a multilayer perceptron, two label prediction heads ($f_\omega(\cdot)$ and $g_\gamma(\cdot)$), a structural self-contrasting network, as well as two optimization losses.}
    \vspace{-0.5em}
	\label{fig:3}
\end{figure*}

\subsection{Structural Self-Contrasting Network} \label{sec:3.3}
Not involving any explicit message passing, the structural self-contrasting network treats the structural information implicitly as prior to guide the computation of supervision signals. In this section, we introduce the structural self-contrasting network from the following three aspects: (1) backbone architecture design, including an MLP and two prediction heads; (2) objective function design, how to self-contrast between the target node $v_i$ and its neighboring node $v_j\in\mathcal{V}_i$; (3) optimization difficulty and strategy, including how to properly sample negative samples and support batch-style training.  A high-level overview of the proposed GSSC framework is shown in Fig.~\ref{fig:3}.

\subsubsection{\textbf{Architecture}}
The structural self-contrasting network is based on a pure MLP architecture, with each layer composed of a linear transformation, an activation function, a batch normalization, and a dropout function, formulated as:
\begin{equation}
\mathbf{H}^{(l)}=\operatorname{Dropout}\big(B N\big(\sigma\big(\mathbf{H}^{(l-1)} \mathbf{W}^{(l-1)}\big)\big)\big),\quad \mathbf{H}^{(0)}=\mathbf{X}
\end{equation}
where $1 \leq l \leq L$, $\sigma\!=\!\mathrm{ReLu}(\cdot)$ denotes an activation function, $BN(\cdot)$ is the batch normalization, and $\operatorname{Dropout}(\cdot)$ is the dropout function. $\mathbf{W}^{(0)} \in \mathbb{R}^{d \times F}$ and $\mathbf{W}^{(l)} \in \mathbb{R}^{F \times F}$ $(1 \leq l \leq L-1)$ are layer-specific weight matrices with the hidden dimension $F$. Furtheromore, we define two additional prediction heads: $\mathbf{y}_i=f_\omega(\mathbf{h}_i^{(L)}) \in \mathbb{R}^{C}$ and $\mathbf{z}_j=g_\gamma(\mathbf{h}_j^{(L)}) \in \mathbb{R}^{C}$, where $C$ is the number of categories.

\subsubsection{\textbf{Structural Smoothness Constraint}} \label{sec:3.3.2}
The structural smoothness assumption indicates that connected nodes should be similar, while disconnected nodes should be far away. With such motivation, we propose a structural smoothness constraint that enables the model to learn the graph connectivity and disconnectivity without explicit message passing. Given a target node $v_i$, we first generate an augmented node $v_{i \rightarrow j}$ between node $v_i$ and its neighboring node $v_j\in\mathcal{N}_i$ by \emph{learnable} interpolation, with its node representation $\mathbf{g}_{i \rightarrow j}$ defined as
\begin{equation}
\begin{aligned}
\mathbf{g}_{i \rightarrow j} = g_\gamma\big(\beta_{i,j}\mathbf{h}_j^{(L)}+(1-\beta_{i,j})\mathbf{h}_i^{(L)}\big), \\ \text{where}\ \  \beta_{i,j}=\mathrm{sigmoid}\big(\mathbf{a}^T \big[\mathbf{h}_i^{(L)}\|\mathbf{h}_j^{(L)}\big]\big)
\end{aligned}
\end{equation}
where $\beta_{i,j}$ is defined as \textit{learnable interpolation coefficients} with the shared weight $\mathbf{a}$. Then, we take the generated augmented node $v_{i \rightarrow j}$ as a positive sample and other non-neighboring nodes as negative samples and define the constraint between nodes $v_i$, $v_j$ as follows
\begin{equation}
\hspace{-1em}
\begin{small}
\begin{aligned}
l_{i,j}  & =  \log \frac{ e^{\mathcal{D}\left(\mathbf{y}_{i}, \mathbf{g}_{i \rightarrow j}\right)}}{\sum_{e_{i,k}\notin\mathcal{E}_g^{\prime}} e^{\mathcal{D}\left(\mathbf{y}_{i}, \mathbf{z}_{k}\right)}} \\ & =  \mathcal{D}\left(\mathbf{y}_{i}, \mathbf{g}_{i \rightarrow j}\right) - \log \sum_{e_{i,k}\notin\mathcal{E}_g^{\prime}} e^{\mathcal{D}\left(\mathbf{y}_{i}, \mathbf{z}_{k}\right)},
\end{aligned}
\end{small}
\end{equation}
where $\mathbf{y}_i=f_\omega(\mathbf{h}_i^{(L)})$, $\mathbf{z}_k=g_\gamma(\mathbf{h}_k^{(L)})$, and $\mathcal{D}: \mathbb{R}^{C} \times \mathbb{R}^{C} \rightarrow \mathbb{R}$ is a discriminator that maps two representations to an agreement score and taken as Mean Square Error (MSE) in our implementation by default. The motivations why we adopt learnable interpolation to augment nodes is based on the following judgment: compared with those non-neighboring nodes, the number of neighboring nodes is much smaller, which makes the model overemphasize the differences between the target and non-neighboring nodes, resulting in imprecise class boundaries. We have demonstrated the benefits of node augmentation and negative samples in Table.~\ref{tab:1}.

The total structural smoothness constraints over the edge set $\mathcal{E}_g^{\prime}$ of the sparsified subgraph $g'$ can be defined as
\begin{equation}
\begin{small}
\begin{aligned}
\mathcal{L}_{smooth} = \frac{1}{N}\sum_{i=1}^N \sum_{e_{i,j}\in\mathcal{E}_g^{\prime}} \Big( \mathcal{D}\left(\mathbf{y}_{i}, \mathbf{g}_{i \rightarrow j}\right) - \log \sum_{e_{i,k}\notin\mathcal{E}_g^{\prime}} e^{\mathcal{D}\left(\mathbf{y}_{i}, \mathbf{z}_{k}\right)} \Big)
\label{equ:10}.
\end{aligned}
\end{small}
\end{equation}

\subsubsection{\textbf{Optimization Difficulty and Strategy}}
Directly optimizing Eq.~(\ref{equ:10}) is computationally expensive for two tricky optimization difficulties: \textit{(1)} it treats all non-neighboring nodes as negative samples, which suffers from both data redundancy and huge computational burden; and \textit{(2)} it performs the summation over the entire set of nodes, i.e, requiring a large memory space for keeping the entire graph. To address these problems, we adopt the edge sampling strategy \citep{mikolov2013distributed} for batch-style training. More specifically, we first sample a mini-batch of edges from the entire edge set $\mathcal{E}_g^{\prime}$ to construct a mini-batch $\mathcal{E}_b\in\mathcal{E}_g^{\prime}$. Then we randomly sample negative nodes from a pre-defined negative sample distribution $P_k(v)$ for each edge $e_{i,j}\in\mathcal{E}_b$ instead of enumerating all non-neighboring nodes as negative samples. Finally, we can rewrite Eq.~(\ref{equ:10}) as follows
\begin{equation}
\begin{small}
\begin{aligned}
\mathcal{L}_{smooth}  = & \frac{1}{B}\sum_{b=1}^B  \sum_{e_{i,j}\in\mathcal{E}_b}   \Big(
\mathcal{D}\left(\mathbf{y}_{i}, \mathbf{g}_{i \rightarrow j}\right) \!+\! \mathcal{D}\left(\mathbf{y}_{j}, \mathbf{g}_{j \rightarrow i}\right) \\ & - \mathbb{E}_{{v_k} \sim P_k(v)} \big(\log e^{\mathcal{D}\left(\mathbf{y}_{i}, \mathbf{z}_{k}\right)} + \log e^{\mathcal{D}\left(\mathbf{y}_{j}, \mathbf{z}_{k}\right)}\big)
\Big),
\end{aligned}
\end{small}
\label{equ:11}
\end{equation}
where $B$ is the batch size, and $v_k$ is a random sample drawn from the pre-defined negative sample distribution $P_k(v_i)=\frac{d_i}{|\mathcal{E}_g^{\prime}|}$ for each node $v_i$, where $d_i$ is the degree of node $v_i$.

Similarly, we can formulate the cross-entropy loss on the labeled node set $\mathcal{V}_L$ as a classification loss, as follows
\begin{equation}
\hspace{-1em}
\mathcal{L}_{cla}=\frac{1}{B}\sum_{b=1}^B\sum_{i\in\mathcal{V}_L\cap\mathcal{V}_b}\Big(CE(y_i, \widehat{\mathbf{y}}_{i}) + \sum_{e_{i,j}\in\mathcal{E}_b}
CE(y_i, \widehat{\mathbf{z}}_{j})\Big),
\end{equation}
where $\mathcal{V}_b=\{v_i,v_j|e_{i,j}\in\mathcal{E}_b\}$ is all the sampled nodes in $\mathcal{E}_b$, $\widehat{\mathbf{y}}_{i}=\textrm{softmax}(\mathbf{y}_i) \in \mathbb{R}^C$, and $\widehat{\mathbf{z}}_{j}=\textrm{softmax}(\mathbf{z}_j) \in \mathbb{R}^C$. Besides, $CE(\cdot)$ is the cross-entropy loss, and $y_i$ is the ground-truth label of node $v_i$. The total training loss is defined as:
\begin{equation}
\mathcal{L}_{total}\left(g^\prime(\psi),\theta\right)=\mathcal{L}_{smooth}+\mathcal{L}_{cla}.
\label{equ:13}
\end{equation}
\textcolor{new}{Note that Eq.~(\ref{equ:13}) is defined for node classification and needs some minor modifications to be extended to graph-level tasks. Since there are no node labels in graph-level tasks, it requires replacing $\mathcal{L}_{cla}$ in Eq.~(\ref{equ:13}) with another loss $\mathcal{L}_{graph}$. Therefore, we aggregate the node embeddings in a graph into a graph embedding by average pooling, and then compute the loss between it and the graph label, e.g., MSE for a regression task, or cross-entropy for a classification task. The total training loss for graph-level tasks is $\mathcal{L}_{total}\left(g^\prime(\psi),\theta\right)=\mathcal{L}_{smooth}+\mathcal{L}_{graph}$.}

\subsection{Optimization Strategy} \label{sec:3.4}

\subsubsection{\textbf{Problem Statement}} 
Since the structural information $g^\prime$ is not explicitly involved in the forward propagation in the STR-Contrast network in Eq.~(\ref{equ:11}), we have $\frac{\partial \mathcal{L}_{smooth}}{\partial g^\prime(\psi)}=0$, i.e., the parameter $\psi$ of the sparsification network cannot be directly optimized end-to-end through downstream supervision as many existing sparsification methods have done. A very straightforward idea is to directly modify Eq.~(\ref{equ:11}) as follows
\begin{equation}
\begin{small}
\begin{aligned}
\mathcal{L}_{smooth}^\prime  =  & \frac{1}{B}\sum_{b=1}^B   \sum_{e_{i,j}\in\mathcal{E}_b}  g^\prime_{i,j}\Big(
\mathcal{D}\left(\mathbf{y}_{i}, \mathbf{g}_{i \rightarrow j}\right) \!+\! \mathcal{D}\left(\mathbf{y}_{j}, \mathbf{g}_{j \rightarrow i}\right) \\ & - \mathbb{E}_{{v_k} \sim P_k(v)} \big(\log e^{\mathcal{D}\left(\mathbf{y}_{i}, \mathbf{z}_{k}\right)} + \log e^{\mathcal{D}\left(\mathbf{y}_{j}, \mathbf{z}_{k}\right)}\big)
\Big)
\label{equ:14}
\end{aligned}
\end{small}
\end{equation}
where Eq.~(\ref{equ:14}) allows structural information $g^\prime$ to be explicitly involved in the computation of the supervision signal, so the sparsification network can now be directly optimized by loss $\mathcal{L}_{smooth}^\prime(\cdot)$. \textcolor{mark}{However, optimizing Eq.~(\ref{equ:14}) may result in trivial solutions, i.e., $\mathcal{L}_{smooth}^\prime(\cdot)$ reaches a minimum by forcing $g^\prime_{i,j}$ close to 0 rather than minimizing $l_{i,j}$, and thus an empty edge set $\mathcal{E}^\prime_g$ is learned. The experimental results in subsection \ref{sec:4.4} confirm the potential trend towards trivial solutions.} Therefore, we turn to another more sophisticated design by formulating structural sparsification and self-contrasting as a bi-level optimization problem, as follows
\begin{equation}
\max _{\psi} \mathcal{H}\left(g^\prime(\psi), \theta^{*})\right), \text {s.t. } \theta^{*}\!=\!\arg \min _{\theta} \mathcal{L}_{total}\left(g^\prime(\psi),\theta\right)
\label{equ:15}
\end{equation}
where lower-level objective $\mathcal{L}_{total} (g^\prime(\psi), \theta)$ is defined in Eq.~(\ref{equ:13}), and upper-level objective $\mathcal{H}\left(g^\prime(\psi), \theta^{*})\right)$ denotes the quality measure for the sparsified subgraph $g'$. However, as discussed earlier, we cannot directly employ the downstream performance to measure the sparsified subgraph, so we need an unsupervised quality measure $\mathcal{H}(\cdot)$ to evaluate the quality of the sparsified subgraph $g^\prime$. In a nutshell, one of the most critical optimization difficulties is how to construct a proper optimization objective for structural sparsification without the direct access to downstream supervision.

\subsubsection{\textbf{Homophily-oriented Objective}} 
Inspired by the discussions on homophily in Sec.~\ref{sec:3.1}, we design a homophily-oriented objective and used it as a measure for the quality of the sparsified graph. The homophily-oriented objective is defined as follows
\begin{equation}
    \max _{\psi} \mathcal{H}\left(g^\prime(\psi), \theta^{*})\right)=\frac{\sum_i \sum_{j\in\mathcal{E}_g^\prime}g^\prime_{i,j}\cdot\mathbb{I}(s_i=s_j)}{|\mathcal{E}_g^\prime|}
    \label{equ:16}
\end{equation}
where $\mathbb{I}(\cdot)$ is an indicator function, $s_i=\text{argmax}(\mathbf{y}_i)$ and $s_j=\text{argmax}(\mathbf{y}_j)$ are the pseudo-labels obtained from self-contrasting network $q_{\theta^{*}}(Y \mid \mathbf{X}, g^\prime)$. Extensive experiments have been provided in Sec.~\ref{sec:4.4} to demonstrate the effectiveness of the homophily-oriented objective in Eq.~(\ref{equ:16}).

\subsubsection{\textbf{Bi-level optimization}}
We adopt the alternating gradient descent (AGD) algorithm \cite{wang2019towards} to optimize Eq.~(\ref{equ:15}), alternating between upper-level maximization and lower-level minimization, with the pseudo-code outlined in Algorithm~\ref{algo:1}.
\newline 

\vspace{-0.8em}
\noindent \textbf{Updating lower-level $\theta$.} The lower-level minimization follows the conventional gradient descent procedure given the sampled sparsified subgraph $g^\prime(\psi^{(n-1)})$, represented as:
\begin{equation}
\theta^{(n)}=\theta^{(n-1)}-\alpha_{\theta} \nabla_{\theta} \mathcal{L}_{total}\left(g^\prime(\psi^{(n-1)}),\theta^{(n-1)}\right)
\label{equ:17}
\end{equation}

\noindent \textbf{Updating upper-level $\psi$.} After updating parameter $\theta^{(n)}$, we perform the upper-level maximization and update $\psi^{(n-1)}$ as:
\begin{equation}
\psi^{(n)}=\psi^{(n-1)}+\alpha_{\psi} \nabla_{\psi} \mathcal{H}\left(g^\prime(\psi^{(n-1)}),\theta^{(n)}\right)
\label{equ:18}
\end{equation}
where $\alpha_{\theta} \in \mathcal{R}_{>0}$ and $\alpha_{\psi} \in \mathcal{R}_{>0}$ are the learning rates.

\begin{algorithm}[!htbp]
	\caption{AGD for bi-level optimization}
	\label{algo:1}
	\begin{algorithmic}[1]
		\State Initialize sparsification network parameter $\phi^{(0)}$ and neighborhood self-contrasting network parameter $\theta^{(0)}$.
		
		\For{$n$ $\in$ \{1, $\cdots$, $N$\}}
		    \State Generate sparsified subgraph $g^\prime(\psi^{(n-1)})$ by Eq.~(\ref{equ:5}-\ref{equ:6}).
		    \State Compute $\mathcal{L}_{total}\left(g^\prime(\psi^{(n-1)}),\theta^{(n-1)}\right)$ by Eq.~(\ref{equ:13});
		    \State \textbf{Lower-level maximization:} Fix parameter $\psi^{(n-1)}$, \Statex \quad \;  and update parameter $\theta^{(n)}$ by Eq.~(\ref{equ:17}).
		    \State Compute $\mathcal{H}\left(g^\prime(\psi^{(n-1)}),\theta^{(n)}\right)$ by Eq.~(\ref{equ:16});
		    \State \textbf{Upper-level minimization:} Fix parameter $\theta^{(n)}$, and 
		    \Statex \quad \; update parameter $\psi^{(n)}$ by Eq.~(\ref{equ:18}).
		\EndFor
		\State \textbf{return} Predicted labels $\mathcal{Y}_U$ and optimized paramter $\theta^{(N)}$
	\end{algorithmic}
\end{algorithm}

\subsection{Discussion and Comparison}
\subsubsection{\textbf{Discuss on Structural Self-contrasting Network}} 
Different from existing GNNs, such as GCN and GAT, that guide the feature aggregation among neighbors through powerful message passing, the structural self-contrasting network is based purely on a multilayer perceptron where structural information is only implicitly used as prior in the computation of supervision signals, but does not explicitly involve in the forward propagation. \textcolor{mark}{Another research topic that is close to us is graph contrastive learning \cite{wu2021self,liu2022graph}, but we differ from them in the three aspects: \textit{(1)} learning objective, graph contrastive learning aims to learn transferable knowledge from abundant unlabeled data in an \emph{unsupervised} setting and then generalize the learned knowledge to downstream tasks. Instead, the structural self-contrasting network works in a \emph{semi-supervised} setting, i.e., the partial label information is available during training. \textit{(2)} augmentation, graph contrastive learning usually requires multiple types of sophisticated augmentation to obtain different views for contrasting \cite{you2020graph,jiao2020sub,hassani2020contrastive}. However, we augment nodes only by simple linear interpolation. \textit{(3)} We remove those noisy edges by the sparsification network, which can be considered as an \textit{``adaptive"} edge-wise self-contrasting.}

\subsubsection{\textbf{Discuss on the structural sparsification network}} 
Different from existing graph sparsification methods that are either attention-based or gradient-based, our STR-Sparse formulates graph sparsification learning as a two-stage process based on a probabilistic generative model, where \textit{(1) in the first stage}, learning the sparsification distribution and modeling the sparsification strategy; \textit{(2) in the second stage}, sampling sparsified subgraphs from the learned sparsification strategy through Gumbel-Softmax sampling, which transforms the problem from a discrete combinatorial optimization to a continuous subgraph generation problem.

\section{Experiments}
\subsection{Experimental setups}
The experiments aim to answer the following six questions:

\begin{itemize}
    \item[\textbf{(Q1)}] How effective is GSSC compared to those GNNs, graph sparsification, and MLP-based models?
    \item[\textbf{(Q2)}] Is GSSC robust to label noise and structure noise?
    \item[\textbf{(Q3)}] How effective is the homophily-oriented objective?
    \item[\textbf{(Q4)}] Is STR-Sparse applicable to other GNNs and MLP-based models? How does STR-Contrast perform with other existing graph sparsification methods?
    \item[\textbf{(Q5)}] How efficient is the model in terms of inference time?
    \item[\textbf{(Q6)}] How do hyperparameters affect performance?
\end{itemize}

\subsubsection{\textbf{Datasets}} The experiments are conducted on ten real-world datasets. For the four small-scale datasets, namely Cora \cite{sen2008collective}, Citeseer \cite{giles1998citeseer}, Coauthor-CS, and Coauthor-Phy \cite{shchur2018pitfalls}, we follow \cite{kipf2016semi,liu2020towards} to select 20 nodes per class to construct a training set, 500 nodes for validation, and 1000 nodes for testing. For the ogbn-arxiv and ogbn-proteins datasets, we use the public data splits provided by the authors \cite{hu2020open}. For the four heterophily graphs, namely Actor, Squirrel, Chameleon, and Deezer, we follow the data splitting strategy by \cite{wu2024simplifying}.

\begin{table*}[!htbp]
\begin{center}
\caption{Classification accuracy $\pm$ std (\%) of general GNNs, graph sparsification, graph transformer, and MLP-based models, where the metrics marked in \textbf{bold}, \underline{\textit{underline}}, and {\color[rgb]{0.75,0.75,0.75}gray} are the top results for rank 1, 2, and 3 (same for Table.~\ref{tab:4} and Table.~\ref{tab:1}).}
\label{tab:1}
\resizebox{1.0\textwidth}{!}{
\begin{tabular}{clcccccc}

\toprule
\textbf{Type} & \textbf{Method} & \textbf{Cora} & \textbf{Citeseer} & \textbf{Coauthor-CS} & \textbf{Coauthor-Phy} & \textcolor{mark}{\textbf{ogbn-arxiv}} & \textcolor{mark}{\textbf{ogbn-proteins}} \\ \midrule
\multirow{6}{*}{\textbf{GNNs}} & GCN \cite{kipf2016semi} & 81.28$\pm$0.42 & 71.06$\pm$0.44 & 88.66$\pm$0.48 & 92.14$\pm$0.34 & \textcolor{mark}{71.74$\pm$0.29} & \textcolor{mark}{72.51$\pm$0.35} \\
& GAT \cite{velivckovic2017graph} & 83.02$\pm$0.45 & 72.56$\pm$0.51 & 89.28$\pm$0.63 & 92.40$\pm$0.52 & \textbf{73.65$\pm$0.11} & 73.44$\pm$0.41 \\
& GraphSAGE \cite{hamilton2017inductive} & 82.22$\pm$0.80 & 71.22$\pm$0.58 & 89.18$\pm$0.45 & 91.54$\pm$0.54 & \textcolor{mark}{69.83$\pm$0.25} & \textcolor{mark}{72.81$\pm$0.46} \\
& SGC \cite{wu2019simplifying} & 80.88$\pm$0.47 & 71.84$\pm$0.72 & 88.56$\pm$0.60 & 90.92$\pm$0.62 & \textcolor{mark}{67.79$\pm$0.27} & \textcolor{mark}{70.31$\pm$0.23} \\
& APPNP \cite{klicpera2018predict} & 83.28$\pm$0.33 & 71.74$\pm$0.27 & 89.72$\pm$0.59 & 92.54$\pm$0.59 & \textcolor{mark}{71.14$\pm$0.28} & \textcolor{mark}{73.17$\pm$0.40} \\
& DAGNN \cite{liu2020towards} & \cellcolor{gray!20}84.30$\pm$0.51 & \cellcolor{gray!20}73.14$\pm$0.62 & 90.20$\pm$0.61 & 93.02$\pm$0.72 & \textcolor{mark}{71.91$\pm$0.23} & \textcolor{mark}{73.75$\pm$0.36} \\ \midrule

\multirow{5}{*}{\textbf{Graph Sparsification}} & SS/RD \cite{sadhanala2016graph,voudigari2016rank} & 79.42$\pm$0.64 & 70.43$\pm$0.52 & 87.89$\pm$0.64 & 91.75$\pm$0.47 & \textcolor{mark}{67.27$\pm$0.23} & \textcolor{mark}{72.13$\pm$0.45} \\
& DropEdge \cite{rong2019dropedge} & 82.23$\pm$0.51 & 72.14$\pm$0.63 & 88.95$\pm$0.57 & 92.43$\pm$0.63 & \textcolor{mark}{71.84$\pm$0.24} & \textcolor{mark}{72.78$\pm$0.54} \\
& LDS \cite{franceschi2019learning} & 83.14$\pm$0.56 & 72.34$\pm$0.70 & 89.24$\pm$0.48 & 92.72$\pm$0.38 & - & - \\
& NeuralSparse \cite{zheng2020robust} & 83.78$\pm$0.54 & \underline{73.19$\pm$0.56} & 89.63$\pm$0.50 & 92.88$\pm$0.43 & \textcolor{mark}{71.45$\pm$0.26} & \textcolor{mark}{73.54$\pm$0.39} \\
& GAUG \cite{zhao2020data} & 83.53$\pm$0.38 & 72.86$\pm$0.33 & 89.90$\pm$0.45 & 93.23$\pm$0.36 & \textcolor{mark}{72.12$\pm$0.25} & \textcolor{mark}{74.13$\pm$0.51} \\ \midrule

\multirow{2}{*}{\textcolor{mark}{\textbf{Graph Transformer}}} & \textcolor{mark}{NAGformer \cite{chen2022nagphormer}} & \textcolor{mark}{84.24$\pm$0.68} & \textcolor{mark}{72.45$\pm$0.46} & \textcolor{mark}{89.37$\pm$0.46} & \textcolor{mark}{92.71$\pm$0.79} & \textcolor{mark}{70.10$\pm$0.28} & \textcolor{mark}{73.64$\pm$0.71} \\
& \textcolor{mark}{SGFormer \cite{wu2024simplifying}} & \textcolor{mark}{\underline{84.50$\pm$0.90}} & \textcolor{mark}{72.60$\pm$0.20} & \textcolor{mark}{\underline{90.42$\pm$0.51}} & \textcolor{mark}{\underline{93.46$\pm$0.74}} & \textcolor{mark}{72.63$\pm$0.13} & \textcolor{mark}{\textbf{79.53$\pm$0.38}} \\ \midrule

\multirow{7}{*}{\textbf{MLP-based Model}} & MLP & 61.86$\pm$0.43 & 59.76$\pm$0.51 & 83.12$\pm$0.53 & 86.24$\pm$0.66 & \textcolor{mark}{55.50$\pm$0.23} & \textcolor{mark}{72.04$\pm$0.48} \\
& Graph-MLP \cite{hu2021graph} & 81.45$\pm$0.52 & 72.87$\pm$0.70 & 89.80$\pm$0.68 & 91.85$\pm$0.49 & OOM & OOM \\
& LinkDist \cite{luo2021distilling} & 76.70$\pm$0.47 & 65.19$\pm$0.55 & 89.56$\pm$0.58 & 92.36$\pm$0.70 & \textcolor{mark}{69.24$\pm$0.26} & \textcolor{mark}{72.45$\pm$0.42} \\
& GSSC (ours) & \textbf{85.28$\pm$0.42} & \textbf{73.74$\pm$0.48} & \textbf{91.00$\pm$0.40} & \textbf{94.40$\pm$0.86} & \textcolor{mark}{\underline{72.90$\pm$0.25}} & \textcolor{mark}{\underline{79.25$\pm$0.31}} \\

& \quad w/o Augmentation & 84.12$\pm$0.48 & 73.10$\pm$0.53 & \cellcolor{gray!20}90.27$\pm$0.51 & \cellcolor{gray!20}93.39$\pm$0.90 & \textcolor{mark}{\cellcolor{gray!20}72.70$\pm$0.24} & \textcolor{mark}{\cellcolor{gray!20}78.90$\pm$0.35} \\
& \quad w/o Negative Samples & 83.42$\pm$0.63 & 72.64$\pm$0.55 & 89.80$\pm$0.62 & 92.76$\pm$0.78 & \textcolor{mark}{72.26$\pm$0.22} & \textcolor{mark}{78.34$\pm$0.42} \\
\bottomrule

\end{tabular}}
\end{center}
\end{table*}

\subsubsection{\textbf{Baseline}} We consider the following six general GNN models: GCN \cite{kipf2016semi}, GAT \cite{velivckovic2017graph}, GraphSAGE \cite{hamilton2017inductive}, SGC \cite{wu2019simplifying}, APPNP \cite{klicpera2018predict}, and DAGNN \cite{liu2020towards}. Besides, we compare the proposed framework with Graph-MLP \cite{hu2021graph} and LinkDist \cite{liu2020towards}, both of which are based on a pure MLP architecture. \textcolor{mark}{Furthermore, two current state-of-the-art graph transformers, SGFormer \cite{wu2024simplifying} and NAGformer \cite{chen2022nagphormer}, are also included in the comparison.} Besides, six graph sparsification methods are included as baselines: Spectral Sparsifier (SS) \cite{sadhanala2016graph}, Rank Degree (RD) \cite{voudigari2016rank}, DropEdge \cite{rong2019dropedge}, LDS \cite{franceschi2019learning}, STR-Sparse \cite{zheng2020robust}, and GAUG \cite{zhao2020data}. When not mentioned specifically, we default to taking the experimental settings of the original paper for implementing these six methods and adopting GCN as the backbone. Each set of experiments is run five times with different random seeds, and the averages and standard deviations are reported. 

\subsubsection{\textbf{Hyperparameter}} The following hyperparameters are set the same for all datasets: Adam optimizer with learning rate $\alpha_\theta$ = $\alpha_\psi$ = 0.01 and weight decay $decay$ = 5e-4; Epoch $E$ = 200; Layer number $L$ = 2. The other dataset-specific hyperparameters are determined by an AutoML toolkit NNI with the search spaces as: hidden dimension $F=\{256, 512, 1024\}$; batch size $B=\{512, 1024, 4096\}$, fusion factor $\alpha=\{0.1, 0.3, 0.5\}$; temperature $\tau=\{0.1, 0.5, 0.8, 1.0\}$. Moreover, we follow \cite{zhao2020data} to first warm-up the structural self-contrasting network with the loss $\mathcal{L}_{total}$ for 100 epochs in the original graph $\mathcal{G}$. 

\begin{table}[!tbp]
\begin{center}
\caption{\textcolor{mark}{Classification accuracy ± std (\%) on 4 heterophily graphs, where the results of various baselines come from \cite{wu2024simplifying}.}}
\label{tab:2}
\resizebox{\columnwidth}{!}{
\begin{tabular}{l|cccc}

\toprule
\textbf{Method} & \textbf{Actor} & \textbf{Squirrel} & \textbf{Chameleon} & \textbf{Deezer} \\ \midrule
GCN & 30.1$\pm$0.2 & 38.6$\pm$1.8 & 41.3$\pm$3.0 & 62.7$\pm$0.7 \\
GAT & 29.8$\pm$0.6 & 35.6$\pm$2.1 & 39.2$\pm$3.1 & 61.7$\pm$0.8 \\
SGC & 27.0$\pm$0.2 & 39.3$\pm$2.3 & 39.0$\pm$3.3 & 62.3$\pm$0.4 \\
APPNP & 31.1$\pm$1.5 & 35.3$\pm$1.9 & 38.4$\pm$3.5 & 66.1$\pm$0.6 \\ \midrule
H2GCN \cite{zhu2020beyond} & 34.4$\pm$1.7 & 35.1$\pm$1.2 & 38.1$\pm$4.0 & 66.2$\pm$0.8 \\
SIGN \cite{frasca2020sign} & 36.5$\pm$1.0 & 40.7$\pm$2.5 & 41.7$\pm$2.2 & 66.3$\pm$0.3 \\
GloGNN \cite{li2022finding} & 36.4$\pm$1.6 & 35.7$\pm$1.3 & 40.2$\pm$3.9 & 65.8$\pm$0.8 \\
SGFormer \cite{wu2024simplifying} & \textbf{37.9$\pm$1.1} & \underline{41.8$\pm$2.2} & \underline{44.9$\pm$3.9} & \underline{67.1$\pm$1.1} \\
GSSC (ours) & \underline{37.5$\pm$0.5} & \textbf{43.1$\pm$1.5} & \textbf{45.7$\pm$3.6} & \textbf{67.6$\pm$0.9} \\ \bottomrule

\end{tabular}}
\end{center}
\end{table}

\subsection{Performance for Node\&Graph Classification \textbf{(Q1)}} \label{sec:4.2}
To answer \textit{Q1}, we conduct experiments on six real-world datasets in comparison to other baselines, where SS/RD is the combinational method of Spectral Sparsifier (SS) and Rank Degree (RD). It can be seen from Table.~\ref{tab:1} that (1) While Graph-MLP and LinkDist can achieve comparable performance with GCN on a few datasets, they still lag far behind the state-of-the-art GNN models, such as APPNP and DAGNN, and cannot even match the performance of GraphSAGE and GAT on some datasets. (2) More importantly, the performance of Graph-MLP and LinkDist is hardly comparable to that of graph sparsification methods except for SS/RD, which indicates that these MLP-based models have no advantage in terms of classification performance on clean data. (3) \textcolor{mark}{Instead, our model consistently achieves the best overall performance on 5 of 6 datasets, significantly outperforming other GNN models and graph sparsification methods, and is even comparable to two state-of-the-art graph transformers. For example, our model obtains the best performance on two large-scale ogbn-arxiv and ogbn-proteins datasets, and more notably, it outperforms DAGNN by 0.99\% and 5.50\%, respectively.} 

\begin{table}[!htbp]
\begin{center}
\caption{\textcolor{mark}{Classification accuracy ± std (\%) on 4 graphs for molecular property prediction, where all baselines use GCN as the backbone by default, and their results come from \cite{fan2024decoupling}.}}
\label{tab:3}
\resizebox{\columnwidth}{!}{
\begin{tabular}{l|cccc}

\toprule
\textbf{Method} & \textbf{BBBP} & \textbf{Tox21} & \textbf{MUV} & \textbf{HIV} \\ \midrule
GPT-GNN \cite{hu2020gpt} & 65.3$\pm$1.5 & 74.3$\pm$0.7 & \underline{75.6$\pm$1.8} & 74.8$\pm$1.4 \\
GraphCL \cite{You2020GraphCL} & 69.9$\pm$1.6 & \underline{75.1$\pm$0.8} & 75.1$\pm$1.5 & 74.5$\pm$0.6 \\
GraphLoG \cite{xu2021self} & 66.4$\pm$2.8 & 73.9$\pm$1.4 & 73.5$\pm$1.0 & 75.5$\pm$0.5 \\

G-Motif \cite{rong2020self} & \underline{70.3$\pm$1.5} & 73.1$\pm$0.4 & 74.5$\pm$0.6 & 76.0$\pm$1.3 \\
GraphMAE \cite{hou2022graphmae} & 66.3$\pm$0.7 & 68.8$\pm$0.5 & 71.8$\pm$0.3 & 73.5$\pm$0.8 \\
SimGRACE \cite{xia2022simgrace} & 66.7$\pm$0.6 & 74.3$\pm$0.2 & 74.3$\pm$0.8 & 74.4$\pm$1.0 \\
GraphMVP \cite{liu2022pretraining} & 67.9$\pm$1.0 & 74.6$\pm$0.3 & 75.5$\pm$1.6 & \underline{76.2$\pm$0.8} \\
GSSC (ours) & \textbf{70.8$\pm$0.7} & \textbf{75.8$\pm$0.6} & \textbf{76.3$\pm$1.0} & \textbf{77.9$\pm$1.1} \\ \bottomrule

\end{tabular}}
\end{center}
\end{table}

Furthermore, to evaluate the effectiveness of node augmentation and negative samples described in Sec.~\ref{sec:3.3}, we perform the \textbf{ablation study} without negative samples and augmentation (fixing $\beta_{i,j}=1$), respectively. It can be observed from Table.~\ref{tab:1} that both negative samples and augmentation contribute to improving classification performance. In addition, the performance improvements brought by negative samples outweigh the node augmentation across various datasets.

\textcolor{mark}{To evaluate how well GSSC handles heterophily graphs, we compare GSSC with four models specializing in heterophily graphs, i.e., H2GCN \cite{zhu2020beyond}, SIGN \cite{frasca2020sign}, GloGNN \cite{li2022finding}, and SGFormer \cite{wu2024simplifying}. The experimental results in Table.~\ref{tab:2} show that GSSC achieves the best performance on 3 of the 4 datasets, and is second only to SGFormer on the Actor dataset.}

\textcolor{mark}{We further evaluate how well GSSC handles the graph classification task by molecular property prediction. There are four datasets, BBBP, Tox21, MUV, and HIV \cite{hu2020open} are considered. Besides, seven classical graph self-supervised models are included as baselines for comparison. Table.~\ref{tab:3} shows that GSSC outperforms the other baselines on all four datasets, demonstrating the potential of GSSC for graph-level tasks.}

\begin{table*}[!tbp]
\begin{center}
\caption{Classification accuracy $\pm$ std (\%) with different label noise types and ratios.}
\label{tab:4}
\resizebox{1.0\textwidth}{!}{
\begin{tabular}{clccccc|cccc|ccc}

\toprule
\textbf{Dataset} & Flipping-Rate & GCN & GAT & GraphSAGE & APPNP & DAGNN & DropEdge & LDS & NeuralSparse & GAUG & Graph-MLP & LinkDist & GSSC (ours) \\ \midrule
\multirow{8}{*}{\textbf{Cora}} & symmetric 20\% & 77.77$\pm$0.62 & 79.75$\pm$0.92 & 78.38$\pm$0.68 & 79.53$\pm$0.56 & \underline{80.76$\pm$0.59} & 78.16$\pm$0.65 & 78.90$\pm$0.70 & 79.64$\pm$0.62 & \cellcolor{gray!20}79.94$\pm$0.66 & 79.21$\pm$0.64 & 77.21$\pm$0.59 & \textbf{81.36$\pm$0.59} \\
 & symmetric 40\% & 69.39$\pm$0.70 & 72.69$\pm$0.78 & 71.33$\pm$0.81 & 74.38$\pm$0.76 & \cellcolor{gray!20}75.23$\pm$0.86 & 70.48$\pm$0.74 & 72.25$\pm$0.81 & 73.10$\pm$0.93 & 73.38$\pm$0.78 & \underline{75.57$\pm$0.76} & 72.66$\pm$0.81 & \textbf{76.98$\pm$0.40} \\
 & symmetric 60\% & 52.17$\pm$0.93 & 55.16$\pm$0.81 & 53.99$\pm$0.89 & 58.87$\pm$1.04 & \cellcolor{gray!20}63.10$\pm$0.88 & 53.72$\pm$0.75 & 56.45$\pm$0.94 & 57.71$\pm$1.03 & 56.78$\pm$0.77 & \underline{64.35$\pm$0.90} & 57.21$\pm$1.06 & \textbf{67.38$\pm$0.72} \\
 & asymmetric 20\% & 71.97$\pm$0.97 & 73.63$\pm$0.83 & 73.58$\pm$1.05 & 74.53$\pm$0.69 & \underline{76.50$\pm$0.92} & 72.62$\pm$0.97 & 73.80$\pm$1.10 & 74.55$\pm$0.85 & \cellcolor{gray!20}74.78$\pm$0.85 & 74.59$\pm$0.70 & 72.48$\pm$0.84 & \textbf{79.72$\pm$0.69} \\
 & asymmetric 40\% & 64.07$\pm$0.58 & 64.24$\pm$0.78 & 63.86$\pm$0.68 & 65.99$\pm$0.69 & 67.18$\pm$0.66 & 64.59$\pm$0.75 & 66.20$\pm$0.72 & \cellcolor{gray!20}67.36$\pm$0.68 & 66.84$\pm$0.81 & \underline{67.85$\pm$0.94} & 66.31$\pm$0.80 & \textbf{69.40$\pm$1.16} \\
 & asymmetric 60\% & 38.47$\pm$0.95 & 39.38$\pm$0.99 & 39.49$\pm$0.78 & 40.39$\pm$1.01 & \cellcolor{gray!20}42.61$\pm$0.81 & 38.74$\pm$1.14 & 40.18$\pm$0.92 & 40.87$\pm$1.03 & 41.52$\pm$0.88 & \underline{43.54$\pm$1.00} & 41.52$\pm$1.16 & \textbf{47.16$\pm$1.18} \\ \midrule
 
\multirow{8}{*}{\textbf{Citeseer}} & symmetric 20\% & 66.91$\pm$0.58 & 67.61$\pm$0.59 & 67.34$\pm$0.43 & 68.20$\pm$0.48 & \underline{71.25$\pm$0.61} & 67.30$\pm$0.54 & 68.84$\pm$0.68 & 69.22$\pm$0.64 & 69.10$\pm$0.47 & \cellcolor{gray!20}70.53$\pm$0.50 & 62.51$\pm$0.58 & \textbf{72.66$\pm$0.48} \\
 & symmetric 40\% & 61.65$\pm$0.59 & 63.88$\pm$0.46 & 62.21$\pm$0.58 & 65.61$\pm$0.58 & \cellcolor{gray!20}69.32$\pm$0.77 & 62.85$\pm$0.59 & 64.58$\pm$0.65 & 65.84$\pm$0.73 & 65.64$\pm$0.70 & \underline{69.52$\pm$0.69} & 61.23$\pm$0.80 & \textbf{71.68$\pm$0.75} \\
 & symmetric 60\% & 54.83$\pm$0.63 & 55.26$\pm$0.90 & 54.20$\pm$0.58 & 55.84$\pm$0.56 & \cellcolor{gray!20}59.36$\pm$0.77 & 55.05$\pm$0.74 & 57.25$\pm$0.64 & 58.29$\pm$0.69 & 58.46$\pm$0.81 & \underline{61.79$\pm$0.78} & 57.35$\pm$0.73 & \textbf{65.06$\pm$0.82} \\
 & asymmetric 20\% & 65.38$\pm$0.89 & 66.62$\pm$0.85 & 66.52$\pm$0.70 & \cellcolor{gray!20}68.17$\pm$0.96 & \underline{68.61$\pm$0.89} & 65.72$\pm$0.76 & 66.80$\pm$0.81 & 67.42$\pm$0.85 & 67.74$\pm$0.79 & 67.93$\pm$0.95 & 60.62$\pm$0.87 & \textbf{72.74$\pm$0.74} \\
 & asymmetric 40\% & 55.70$\pm$1.07 & 56.42$\pm$0.81 & 56.60$\pm$0.99 & 57.63$\pm$0.79 & \cellcolor{gray!20}60.39$\pm$0.86 & 56.10$\pm$0.77 & 57.86$\pm$0.90 & 58.65$\pm$0.83 & 58.93$\pm$0.70 & \underline{61.76$\pm$0.83} & 54.30$\pm$1.02 & \textbf{67.32$\pm$0.78} \\
 & asymmetric 60\% & 41.90$\pm$0.98 & 43.70$\pm$1.15 & 42.65$\pm$0.58 & 45.15$\pm$0.91 & \cellcolor{gray!20}46.05$\pm$0.87 & 42.21$\pm$0.68 & 43.85$\pm$0.72 & 44.75$\pm$0.74 & 44.48$\pm$0.83 & \underline{48.25$\pm$1.12} & 44.09$\pm$1.20 & \textbf{53.24$\pm$1.09} \\ \bottomrule
 
\end{tabular}}
\end{center}
\end{table*}

\subsection{Evaluation on Robustness \textbf{(Q2)}}
To demonstrate the model's robustness, we evaluate the model under different ratios of (1) label noise and (2) structure perturbations on the Cora and Citeseer datasets.

\subsubsection{\textbf{Performance with Noisy Labels}} The performance with various label noise ratios $r\in\{20\%, 40\%, 60\%\}$ is reported in Table.~\ref{tab:4} for two types of label noise: symmetric and asymmetric \cite{tan2021co,xia2023gnn}. The symmetric noise indicates that label $i$ $(0\leq i \leq C-1)$ of each training sample changes independently with probability $\frac{r}{|C|-1}$ to another class $j$ $(j \neq i)$, but with probability $1-r$ preserved as label $i$; the asymmetric noise indicates that label $i$ flips independently with probability $r$ to fixed class $j$ $(j=(i+1) \% C)$, but with probability $1-r$ preserved as label $i$. It can be seen from Table.~\ref{tab:4} that (1) The Graph-MLP, as a classical MLP-based model, significantly outperforms various GNN models under extremely high label noise, but it slightly falls behind DAGNN and GAUG at low-noise settings. (2) GSSC performs most robustly under various label noise types and ratios, especially with asymmetric noise and high noise ratios. For example, with $r=60\%$ asymmetric noise, our model outperforms DAGNN by 4.55\% and 7.19\% on the Cora and Citeseer datasets, respectively.

\subsubsection{\textbf{Performance with Corrupted Structures}} The classification performance under different structural perturbation ratios $r \in \{10\%, 20\%, 30\%\}$ is reported in Table.~\ref{tab:5}, where the corrupted structures are obtained by randomly removing and adding $r\cdot|\mathcal{E}|$ edges from the original graph for training. It can be seen from Table.~\ref{tab:5} that (1) MLP-based models, such as Graph-MLP, are more advantageous at extremely high ratios of structural perturbations. (2) GSSC is more robust than other baselines under various structural perturbation ratios, especially under severe perturbations. For example, when $r=30\%$, GSSC outperforms NeuralSparse by 1.55\% and 3.06\% on the Cora and Citeseer datasets, respectively.

\textcolor{new}{To further demonstrate the advantages of GSSC in terms of robustness, we select a few baselines that perform well on two large-scale datasets, ogbn-arxiv and obgn-proteins, as shown in Table.~\ref{tab:1}. We compare the performance of GSSC with these baselines under structural perturbations in Table.~\ref{tab:6}. It can be seen that GSSC is only comparable to those traditional GNNs and GTs on clean data without structural noise. However, models that work well on clean data, e.g., GAT and SGFormer, may fail on noisy data; for example, GAT performs poorer than GCN on the obgn-arxiv dataset. By contrast, our GSSC performs best under noisy data, showing superior robustness.}

\begin{table}[!tbp]
\begin{center}
\caption{Classification accuracy $\pm$ std (\%) under structure perturbations on two small-scale datasets, Cora and Citeseer.}
\label{tab:5}
\resizebox{\columnwidth}{!}{
\begin{tabular}{lcccccccc}

\toprule
\multirow{2}{*}{\textbf{Method}} & \multicolumn{3}{c}{\textbf{Cora}} & \multicolumn{3}{c}{\textbf{Citeseer}} \\ \cmidrule(r){2-4} \cmidrule(r){5-7}
 & 10\% & 20\% & 30\% & 10\% & 20\% & 30\% \\ \midrule
GCN & 74.19$\pm$0.96 & 68.69$\pm$0.67 & 63.82$\pm$0.54 & 64.84$\pm$0.69 & 62.87$\pm$0.93 & 60.51$\pm$0.73 \\
GAT & 75.01$\pm$0.97 & 69.62$\pm$0.51 & 64.76$\pm$0.74 & 63.35$\pm$0.89 & 61.81$\pm$0.96 & 58.57$\pm$1.23 \\
GraphSAGE & 74.72$\pm$0.69 & 69.02$\pm$0.50 & 64.14$\pm$0.94 & 63.38$\pm$0.67 & 62.80$\pm$0.66 & 59.54$\pm$0.77 \\
APPNP & 74.36$\pm$0.71 & 70.02$\pm$0.93 & 64.90$\pm$1.05 & 63.88$\pm$0.93 & 61.56$\pm$1.08 & 58.32$\pm$0.69 \\
DAGNN & \cellcolor{gray!20}75.32$\pm$0.85 & 70.41$\pm$0.84 &  \cellcolor{gray!20}65.60$\pm$0.82 & 64.74$\pm$0.65 & 61.92$\pm$0.91 & 58.96$\pm$1.19 \\ \midrule

DropEdge & 74.42$\pm$0.88 & 69.58$\pm$0.75 & 64.12$\pm$0.76 & 65.04$\pm$0.85 & 62.99$\pm$0.95 & 60.89$\pm$0.92 \\
LDS & 74.75$\pm$0.93 & 69.89$\pm$0.71 & 64.47$\pm$0.81 & 65.32$\pm$0.73 & 63.43$\pm$1.10 & 61.13$\pm$1.05 \\
NeuralSparse & 75.23$\pm$0.79 & 70.32$\pm$0.68 & 65.45$\pm$0.72 & \underline{66.10$\pm$0.78} &  \cellcolor{gray!20}63.82$\pm$0.87 & 61.54$\pm$1.30 \\
GAUG & \underline{75.46$\pm$0.90} &  \cellcolor{gray!20}70.73$\pm$0.80 & 64.83$\pm$0.85 & 65.70$\pm$0.80 & 63.65$\pm$0.84 &  \cellcolor{gray!20}61.90$\pm$0.88 \\ \midrule

Graph-MLP & 75.23$\pm$0.74 & \underline{70.90$\pm$0.78} & \underline{65.74$\pm$0.83} &  \cellcolor{gray!20}65.78$\pm$0.80 & \underline{64.20$\pm$1.21} & \underline{62.65$\pm$1.10} \\
LinkDist & 73.92$\pm$0.80 & 69.30$\pm$0.88 & 64.36$\pm$0.98 & 64.55$\pm$0.75 & 61.37$\pm$1.05 & 58.47$\pm$0.94 \\
GSSC (ours) & \textbf{77.42$\pm$0.95} & \textbf{71.98$\pm$0.81} & \textbf{67.00$\pm$0.59} & \textbf{69.70$\pm$0.89} & \textbf{67.06$\pm$0.84} & \textbf{64.60$\pm$0.68} \\ \bottomrule

\end{tabular}}
\end{center}
\end{table}

\begin{table}[!tbp]
\begin{center}
\caption{\textcolor{new}{Classification accuracy $\pm$ std (\%) under structure perturbations on two large-scale datasets, arxiv and proteins.}}
\label{tab:6}
\resizebox{\columnwidth}{!}{
\begin{tabular}{lcccc}

\toprule
\multirow{2}{*}{\textbf{Method}} & \multicolumn{2}{c}{\textbf{ogbn-arxiv}} & \multicolumn{2}{c}{\textbf{ogbn-proteins}} \\ \cmidrule(r){2-3} \cmidrule(r){4-5}
 & 0\% & 20\% & 0\% & 20\% \\ \midrule
GCN & 71.74$\pm$0.29 & 63.53$\pm$0.34 & 72.51$\pm$0.35 & 64.30$\pm$0.44 \\
GAT & \textbf{73.65$\pm$0.11} & 62.70$\pm$0.23 & 73.44$\pm$0.41 & 65.74$\pm$0.50 \\
GraphSAGE & 69.83$\pm$0.25 & 62.49$\pm$0.41 & 72.81$\pm$0.46 & 64.93$\pm$0.38 \\
DAGNN & 71.91$\pm$0.23 & 63.74$\pm$0.42 & 73.75$\pm$0.36 & 66.45$\pm$0.56 \\ \midrule
NeuralSparse & 71.45$\pm$0.26 & 64.51$\pm$0.36 & 73.54$\pm$0.39 & 67.16$\pm$0.46 \\
GAUG & 72.12$\pm$0.25 & \underline{65.04$\pm$0.21} & 74.13$\pm$0.51 & 67.47$\pm$0.64 \\
NAGformer & 70.10$\pm$0.28 & 63.40$\pm$0.35 & \textbf{73.64$\pm$0.71} & 66.21$\pm$0.51 \\
SGFormer & 72.63$\pm$0.13 & 64.28$\pm$0.38 & 79.53$\pm$0.38 & \underline{69.55$\pm$0.43} \\
GSSC (ours) & \underline{72.90$\pm$0.25} & \textbf{66.25$\pm$0.28} & \underline{79.25$\pm$0.31} & \textbf{70.20$\pm$0.42} \\ \bottomrule

\end{tabular}}
\end{center}
\end{table}

\subsection{Evaluation on Homophily Objective \textbf{(Q3)}} \label{sec:4.4}
To evaluate the effectiveness of homophily-oriented objective $\mathcal{H}(\cdot)$, we conduct a more in-depth analysis through the following two sets of experiments: (1) correlation analysis between the homophily ratio of the sparsified graph and the downstream performance and (2) evolution curves of the edge number, homophily ratio (in the learned sparsified graph), and downstream performance during the training process.

\subsubsection{\textbf{Correlation Analysis}}
\begin{figure}[!htbp]
	\begin{center}
		\includegraphics[width=0.4\linewidth]{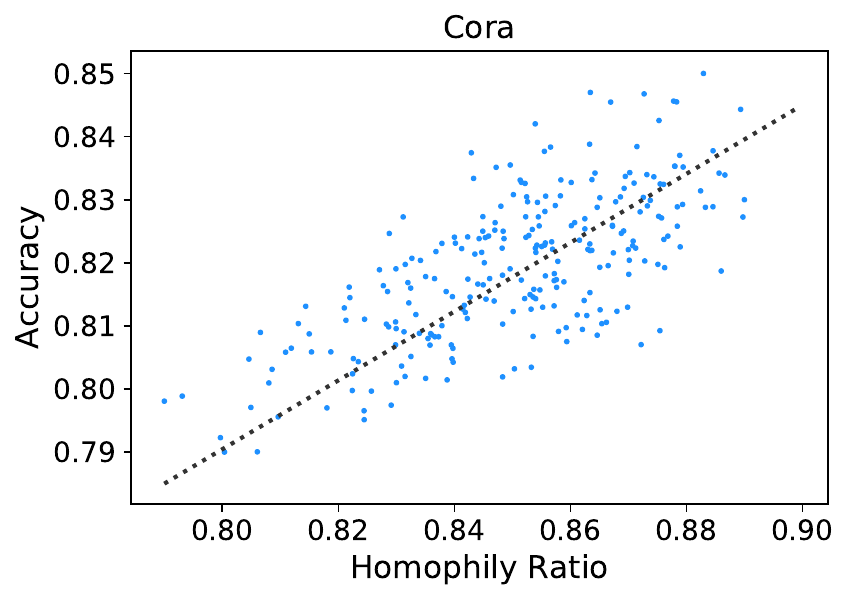}
		\includegraphics[width=0.4\linewidth]{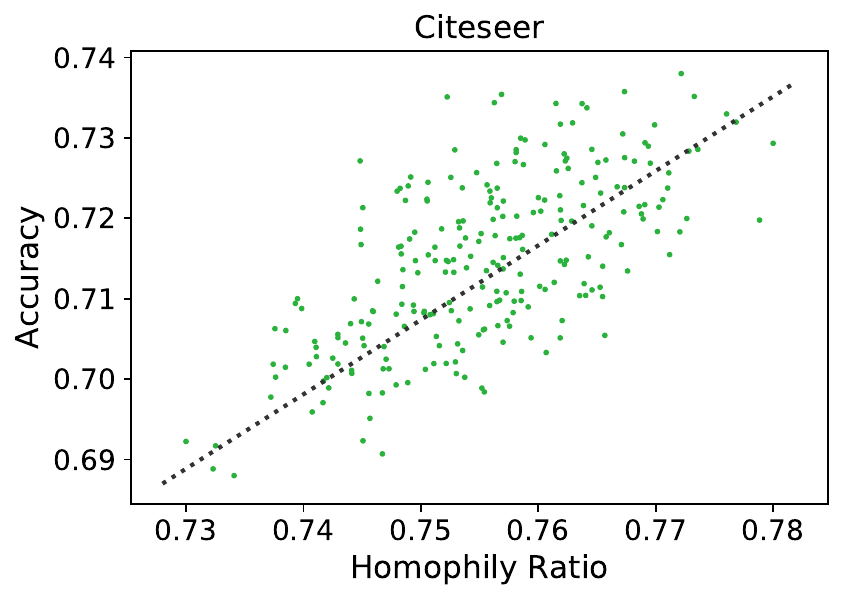}
	\end{center}
	\caption{Correlation between accuracy and homophily ratio.}
	\label{fig:4}
\end{figure}

We randomly remove a fraction of edges from the original graph to obtain a set of candidate subgraphs with different homophily ratios and then train a structural self-contrasting network \textbf{from scratch} on each candidate subgraph. The homophily ratio of the candidate sparsified subgraphs and their corresponding downstream performance are visualized in Fig.~\ref{fig:4}, from which we can observe that the downstream performance tends to be better on the sparsified subgraph with a higher homophily ratio. The tight correlation between homophily ratio and classification accuracy inspires us that homophily ratio may be a desirable option for evaluating the quality of sparsified subgraphs.

\subsubsection{\textbf{Training Evolution Visualization}}
\begin{figure}[!tbp]
	\begin{center}
		\subfigure[Training w/ homophily-oriented objective $\mathcal{H}(\cdot)$]{
		\includegraphics[width=0.49\linewidth]{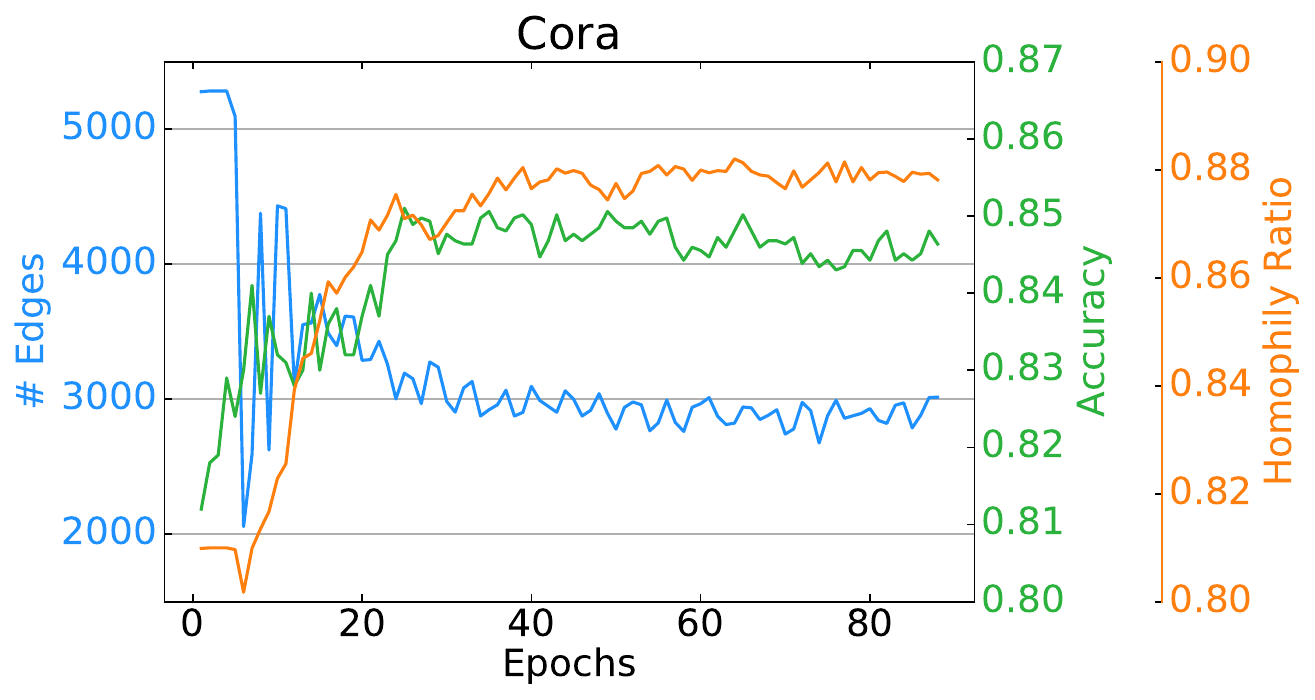}
		\includegraphics[width=0.49\linewidth]{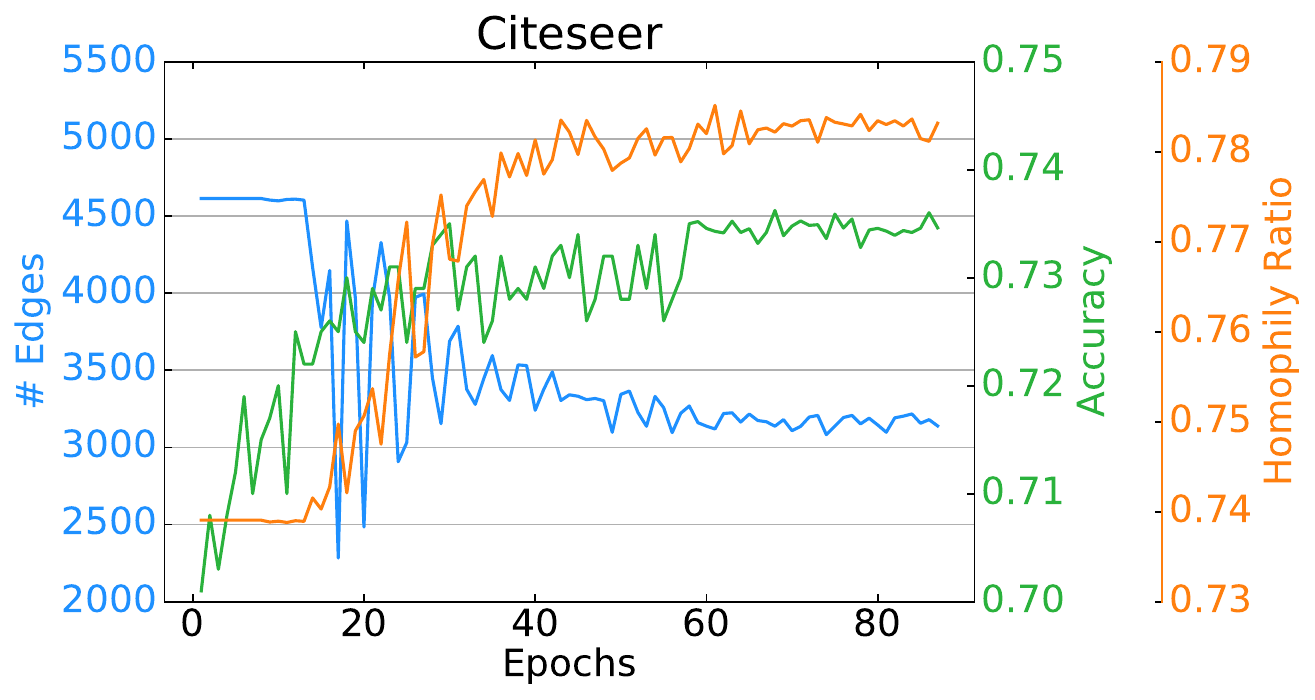}\label{fig:5a}} 
		\subfigure[\textcolor{mark}{Training w/o homophily-oriented objective $\mathcal{H}(\cdot)$}]{
		\includegraphics[width=0.49\linewidth]{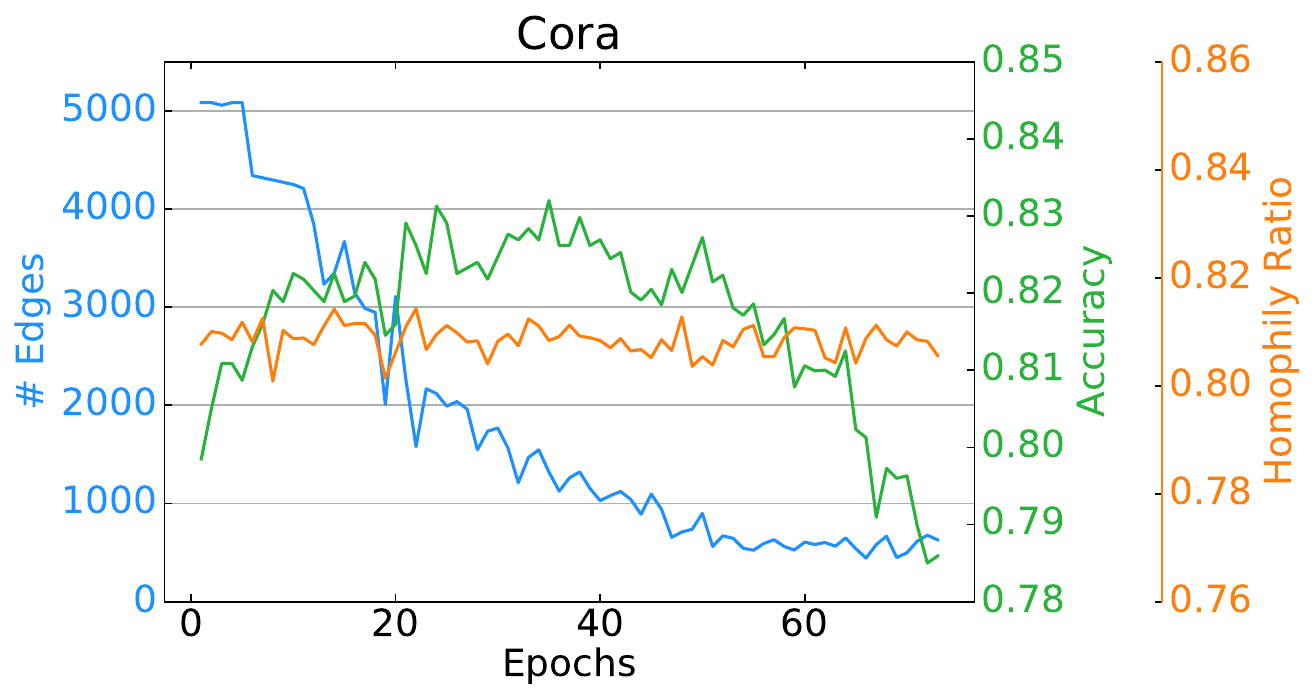}
		\includegraphics[width=0.49\linewidth]{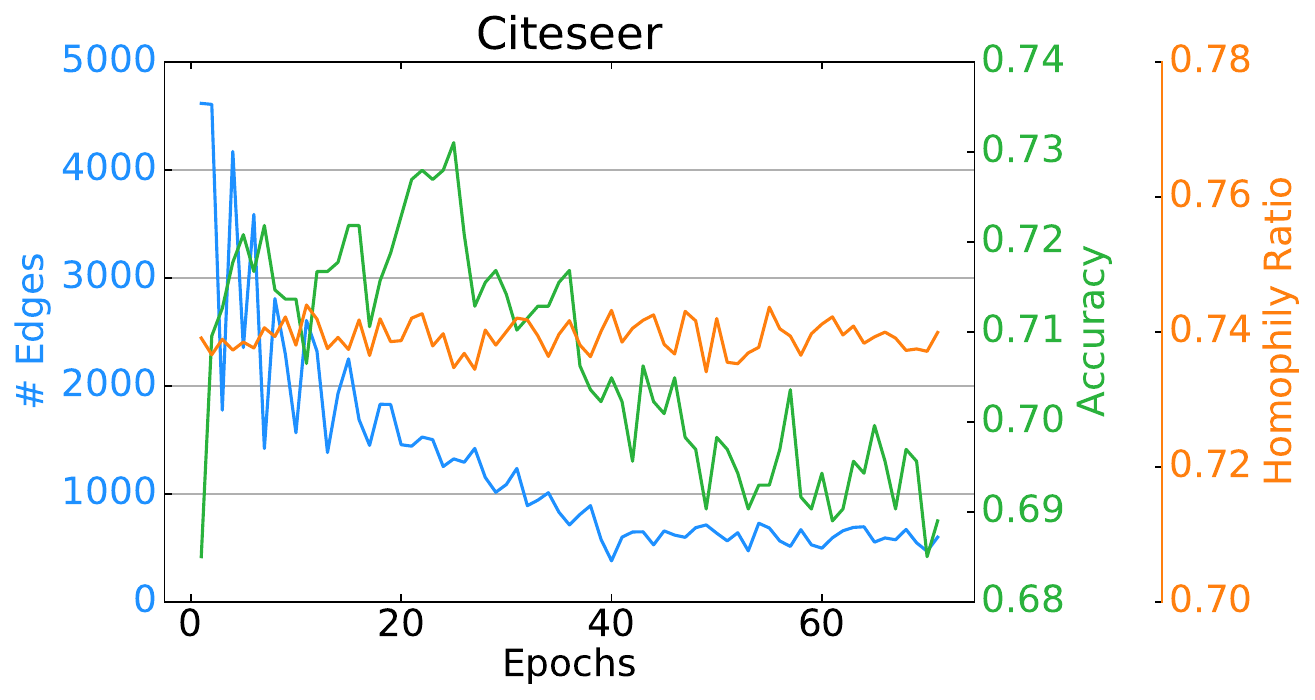}\label{fig:5b}}
	\end{center}
	\caption{Visualization of the training process \textit{with} (Top) and \textit{without} (Bottom) homophily-oriented objective $\mathcal{H}(\cdot)$.}
	\label{fig:5}
\end{figure}

\textcolor{mark}{We visualize in Fig.~\ref{fig:5a} the evolution of edge number, homophily ratio (in the learned sparsified graph), and downstream performance during training. It is clear that the homophily-oriented objective $\mathcal{H}(\cdot)$ can effectively increase the homophily ratio while reducing the number of edges, i.e., making the graph sparse. More importantly, the downstream performance also improves steadily as the homophily ratio increases until it finally converges. To further analyze the role of homophily-oriented objective $\mathcal{H}(\cdot)$, we also train the model with the objective defined in Eq.~(\ref{equ:14}) instead of $\mathcal{H}(\cdot)$ as a comparison. It can be seen that Eq.~(\ref{equ:14}) may result in \emph{trivial solutions, i.e., the number of edges decreases sharply to a small value} with the homophily ratio unchanged as shown in Fig.~\ref{fig:5b}. More importantly, we observe that their downstream performance first improves as the edge number decreases and then drops sharply on the obtained over-sparsified graph, which suggests that over-sparsification is harmful to model performance.}

\subsection{Analysis on STR-Sparse \& STR-Contrast \textbf{(Q4)}}
\subsubsection{\textbf{Evaluation on STR-Sparse}}
Table.~\ref{tab:7} evaluates whether STR-Sparse is applicable to other models, where six GNN models are jointly optimized with STR-Sparse by downstream supervision and three MLP-based models are optimized by the proposed homophily-oriented objective $\mathcal{H}(\cdot)$ in Eq.~\ref{equ:16}. We make the following observations: \textit{(1)} STR-Sparse consistently improves performance across various models compared to the results reported in Table.~\ref{tab:1}; \textit{(2)} MLP-based models benefit more from STR-Sparse than GNNs.

\begin{table}[!htbp]
\begin{center}
\caption{Accuracy $\pm$ std (\%) of GNNs and MLP-based models with Structural Sparsification (STR-Sparse).}
\label{tab:7}
\resizebox{\columnwidth}{!}{
\begin{tabular}{lccccc}

\toprule
\textbf{Model} & \textbf{Cora} & \textbf{Citeseer} & \textbf{Actor} & \textbf{Coauthor-CS} & \textbf{Coauthor-Phy} \\ \midrule
GCN & 81.28$\pm$0.42 & 71.06$\pm$0.44 & 24.84$\pm$0.56 & 88.66$\pm$0.48 & 92.14$\pm$0.34 \\
\ \ + STR-Sparse & 83.45$\pm$0.60 & 72.96$\pm$0.52 & 25.89$\pm$0.70 & 89.52$\pm$0.64 & 92.84$\pm$0.49 \\  \midrule
GAT & 83.02$\pm$0.45 & 72.56$\pm$0.51 & 26.28$\pm$0.45 & 89.28$\pm$0.63 & 92.40$\pm$0.52 \\
\ \ + STR-Sparse  & 83.92$\pm$0.61 & 73.30$\pm$0.68 & 27.88$\pm$0.56 & 89.82$\pm$0.79 & 93.13$\pm$0.65 \\  \midrule
GraphSAGE & 82.22$\pm$0.80 & 71.22$\pm$0.58 & 26.54$\pm$0.70 & 89.18$\pm$0.45 & 91.54$\pm$0.54 \\
\ \ + STR-Sparse  & 83.70$\pm$0.79 & 73.05$\pm$0.63 & 28.05$\pm$0.86 & 90.08$\pm$0.61 & 92.63$\pm$0.69 \\  \midrule
SGC & 80.88$\pm$0.47 & 71.84$\pm$0.72 & 25.24$\pm$0.55 & 88.56$\pm$0.60 & 90.92$\pm$0.62 \\
\ \ + STR-Sparse  & 82.61$\pm$0.54 & 72.62$\pm$0.56 & 26.42$\pm$0.71 & 89.37$\pm$0.66 & 91.90$\pm$0.73 \\  \midrule
APPNP & 83.28$\pm$0.33 & 71.74$\pm$0.27 & 27.82$\pm$1.02 & 89.72$\pm$0.59 & 92.54$\pm$0.59 \\
\ \ + STR-Sparse  & 84.08$\pm$0.51 & 73.18$\pm$0.57 & 28.91$\pm$0.88 & 90.18$\pm$0.64 & 93.40$\pm$0.57 \\  \midrule
DAGNN & 84.30$\pm$0.51 & 73.14$\pm$0.62 & 28.98$\pm$0.86 & 90.20$\pm$0.61 & 93.02$\pm$0.72 \\
\ \ + STR-Sparse  & 84.52$\pm$0.48 & 73.55$\pm$0.74 & 30.19$\pm$0.79 & 90.49$\pm$0.72 & 93.88$\pm$0.76 \\ \midrule \midrule
Graph-MLP & 81.45$\pm$0.52 & 72.87$\pm$0.70 & 25.40$\pm$0.49 & 89.80$\pm$0.68 & 91.85$\pm$0.49 \\
\ \ + STR-Sparse  & 83.50$\pm$0.46 & 73.38$\pm$0.69 & 29.25$\pm$0.75 & 90.22$\pm$0.73 & 93.17$\pm$0.69 \\  \midrule
LinkDist & 76.70$\pm$0.47 & 65.19$\pm$0.55 & 23.96$\pm$0.65 & 89.56$\pm$0.58 & 92.36$\pm$0.70 \\
\ \ + STR-Sparse  & 80.05$\pm$0.55 & 71.20$\pm$0.66 & 27.59$\pm$0.80 & 90.48$\pm$0.64 & 93.46$\pm$0.77 \\  \midrule
GSSC (w/o STR-Sparse) & 83.16$\pm$0.50 & 71.89$\pm$0.78 & 27.96$\pm$0.76 & 89.41$\pm$0.48 & 92.65$\pm$0.49 \\
\ \ + STR-Sparse  & \textbf{85.28$\pm$0.42} & \textbf{73.74$\pm$0.48} & \textbf{31.24$\pm$1.09} & \textbf{91.00$\pm$0.40} & \textbf{94.40$\pm$0.86} \\ \bottomrule

\end{tabular}}
\end{center}
\end{table}

\subsubsection{\textbf{Evaluation on  STR-Contrast}}
Table.~\ref{tab:8} evaluates the applicability of different graph sparsification methods to STR-Contrast, where SS/RD is the combinational method of Spectral Sparsifier (SS) and Rank Degree (RD). The three supervised sparsification methods, LDS, STR-Sparse, and GAUG, are optimized by the homophily-oriented objective $\mathcal{H}(\cdot)$ for a fair comparison. We make the following observations: \textit{(1)} Compared with the vanilla STR-Contrast model (\textbf{without any sparsification, denoted as ``N/A"}), all graph sparsification methods except SS/RD can improve generalization; \textit{(2)} Compared with unsupervised sparsification methods, such as DropEdge, STR-Sparse achieves up to 1.86\% and 2.55\% improvements on the Cora and Actor datasets. Even when compared to supervised sparsification methods, such as STR-Sparse and GAUG, STR-Sparse still shows a huge advantage.

\begin{table}[!tbp]
\begin{center}
\caption{Accuracy $\pm$ std (\%) of the STR-Contrast network with other graph sparsification methods, where "N/A" denotes the results without any sparsification as a baseline.}
\label{tab:8}
\resizebox{\columnwidth}{!}{
\begin{tabular}{lccccc}

\toprule
\textbf{Sparsifier} & \textbf{Cora} & \textbf{Citeseer} & \textbf{Actor} & \textbf{Coauthor-CS} & \textbf{Coauthor-Phy} \\ \midrule
N/A (STR-Contrast) & 83.16$\pm$0.50 & 71.89$\pm$0.78 & 27.96$\pm$0.76 & 89.41$\pm$0.48 & 92.65$\pm$0.49 \\
\ \ + SS/RD & 80.50$\pm$0.76 & 69.74$\pm$0.81 & 24.26$\pm$0.69 & 86.32$\pm$0.62 & 89.58$\pm$0.76 \\
\ \ + DropEdge & 83.42$\pm$0.62 & 72.81$\pm$0.52 & 28.69$\pm$0.54 & 89.73$\pm$0.47 & 93.96$\pm$0.58 \\
\ \ + LDS & 83.56$\pm$0.54 & 72.75$\pm$0.44 & 28.90$\pm$0.62 & 89.84$\pm$0.49 & 93.17$\pm$0.61 \\
\ \ + NeuralSparse & 84.08$\pm$0.42 & 73.22$\pm$0.47 & 29.82$\pm$0.68 & 90.18$\pm$0.55 & 93.44$\pm$0.51 \\
\ \ + GAUG & 84.15$\pm$0.45 & 73.16$\pm$0.39 & 30.37$\pm$0.56 & 89.95$\pm$0.40 & 93.63$\pm$0.43 \\
\ \ + STR-Sparse & \textbf{85.28$\pm$0.42} & \textbf{73.74$\pm$0.48} & \textbf{31.24$\pm$1.09} & \textbf{91.00$\pm$0.40} & \textbf{94.40$\pm$0.86} \\ \bottomrule

\end{tabular}} \vspace{-1em}
\end{center}
\end{table}

\subsection{Evaluation on Complexity Analysis \textbf{(Q5)}} 
\begin{figure}[!htbp]
	\begin{center}
		\includegraphics[width=0.85\linewidth]{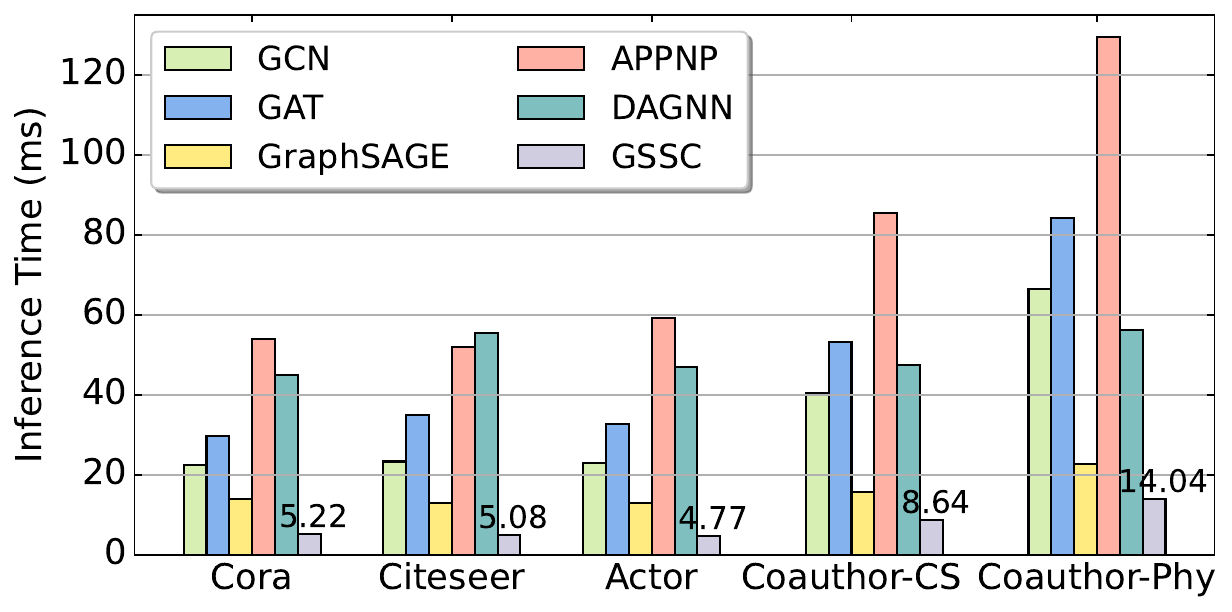}
	\end{center}
	\caption{Inference time \textit{(ms)} for various methods.}
	\label{fig:6}
\end{figure}

The time complexity of the GSSC framework mainly comes from two main parts: (1) STR-Sparse $\mathcal{O}(|\mathcal{V}|dF + |\mathcal{E}|F)$ and (2) STR-Contrast $\mathcal{O}(|\mathcal{V}|dF + |\mathcal{E}^{\prime}_g|F)$, with a total complexity of $\mathcal{O}(|\mathcal{V}|dF+(|\mathcal{E}|+|\mathcal{E}^{\prime}_g|)F)$, where $|\mathcal{E}|$ and $|\mathcal{E}^{\prime}_g|$ are the number of edges in the original and sparsified graph. Due to graph sparsification, we have $|\mathcal{E}^{\prime}_g| \ll |\mathcal{E}|$, so the total complexity is linear w.r.t the number of nodes $|\mathcal{V}|$ and edges $|\mathcal{E}|$, which is in the same order of magnitude as GCN. However, with the removal of neighborhood-fetching latency, the inference time complexity can be reduced from $\mathcal{O}(|\mathcal{V}|dF+|\mathcal{E}|F)$ of GCN to $\mathcal{O}(|\mathcal{V}|dF)$. The inference time ($ms$) averaged over 30 runs is reported in Fig.~\ref{fig:6}, where all methods use $L=2$ layers and hidden dimension $F=16$. Besides, all baselines are implemented based on the standard implementation in the DGL library \citep{wang2019dgl} using the PyTorch 1.6.0 library on NVIDIA v100 GPU. In a fair comparison, we observe that GSSC achieves the fastest inference speed on all datasets.

\subsection{Hyperparameter Sensitivity Analysis (Q6)} 
We evaluate the hyperparameter sensitivity with respect to two key hyperparameters: fusion factor $\alpha$ and batch size $B$, and the results are reported in Fig.~\ref{fig:7} and Fig.~\ref{fig:8}, respectively. In practice, we can determine $\alpha$ and $B$ by selecting the model with the highest accuracy on the validation set.

\subsubsection{Fusion Factor}
\begin{figure}[!htbp]
	\begin{center}
		\includegraphics[width=0.325\linewidth]{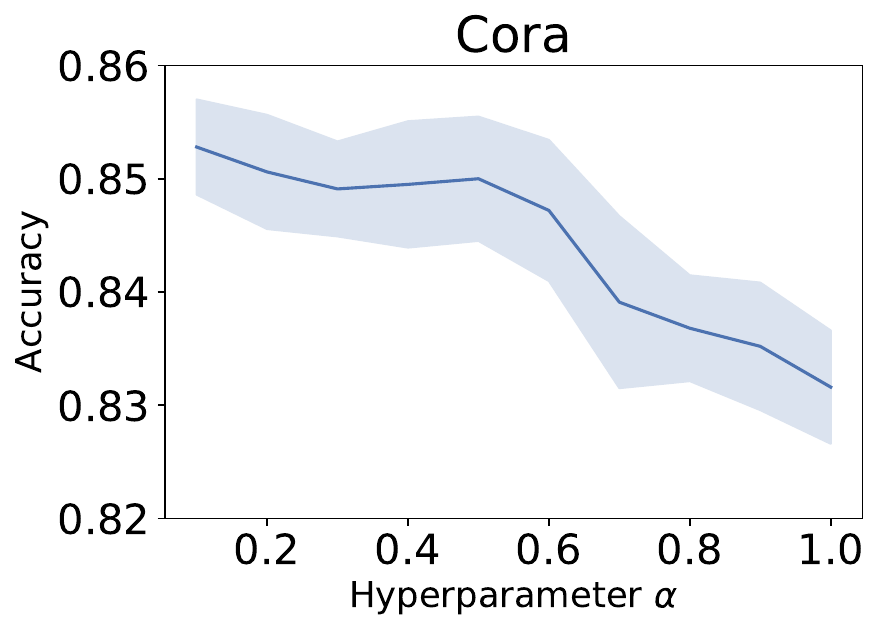}
		\includegraphics[width=0.325\linewidth]{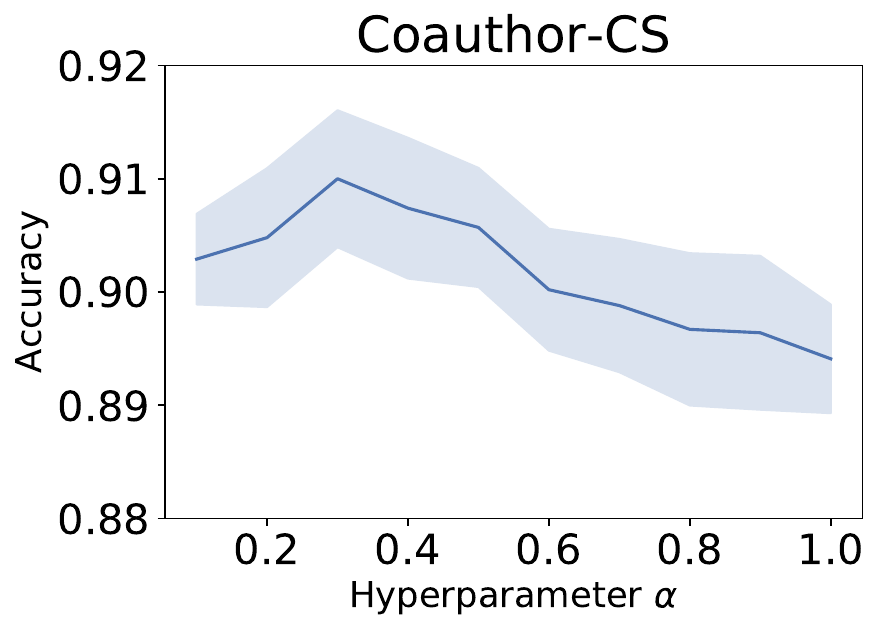}
		\includegraphics[width=0.325\linewidth]{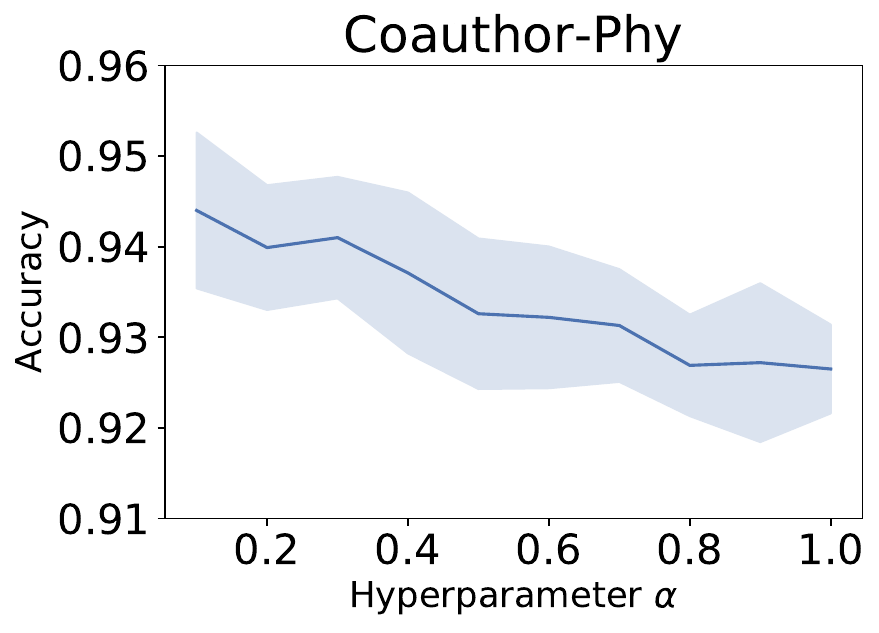}
	\end{center}
	\caption{Parameter sensitivity analysis on fusion factor $\alpha$.}
	\label{fig:7}
\end{figure}

As can be observed in Fig.~\ref{fig:7}, fusion factor $\alpha$ is crucial for the proposed framework. If we set $\alpha=1$, i.e., removing the structural sparsification, the performance is usually the poorest compared with other settings. In practice, we find that setting $\alpha$ to a small value, e.g., $\alpha=0.1$ usually produces pretty good performance. However, a too-small $\alpha$ may cause the sparsified subgraph to deviate too much from the original graph on some datasets, resulting in poor performance. For example, on the Coauthor-CS dataset, the model achieves much better performance when setting $\alpha$ as 0.3 than 0.1. In our experiments, we have only tested with the smallest $\alpha=0.1$, and further performance improvements are expected by finer-grained hyperparameter search.

\subsubsection{\textbf{Batch Size}}
\begin{figure}[!htbp]
	\begin{center}
		\includegraphics[width=0.8\linewidth]{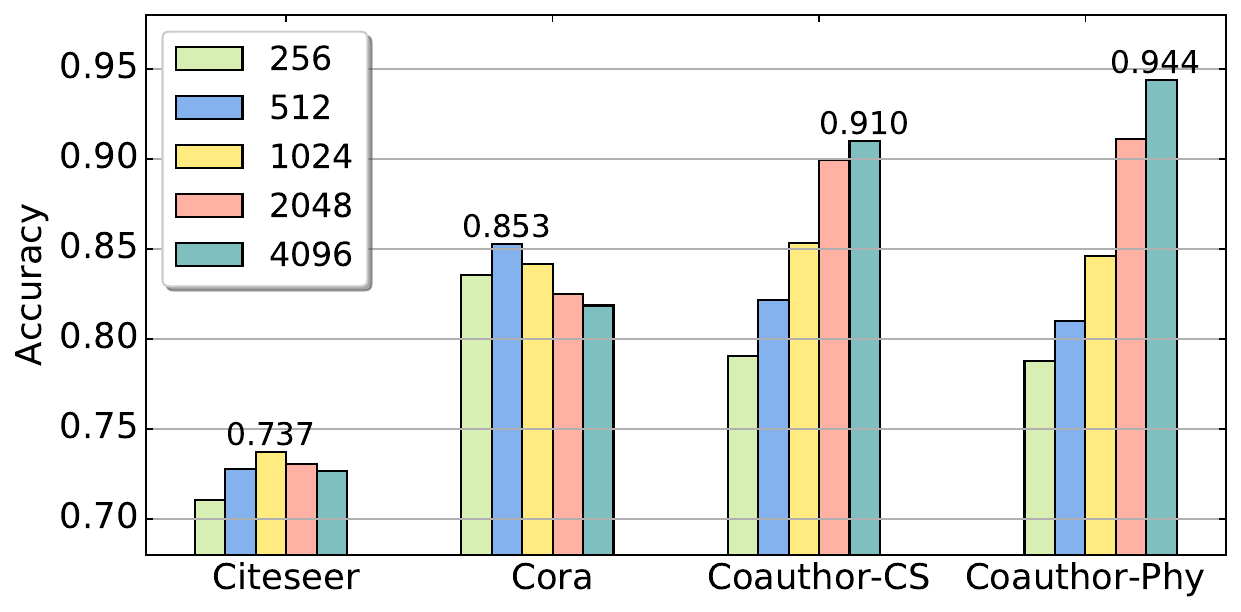}
	\end{center}
	\caption{Parameter sensitivity analysis on batch size $B$.}
	\label{fig:8}
\end{figure}

Fig.~\ref{fig:8} shows the performance of GSSC trained with batch size $B\in\{\text{256, 512, 1024, 2048, 4096}\}$, from which we observe that $B$ is a dataset-specific hyperparameter. For simple graphs with few nodes and edges, such as Cora and Citeseer, a small batch size, $B=\text{256 or 512}$, can yield fairly good performance. However, for large-scale graphs, such as Coauthor-CS and Coauthor-Phy, the performance improves with the increase of batch size $B$.  

\section{Conclusion}
In this paper, we propose a novel \textit{\underline{G}raph \underline{S}tructural \underline{S}elf-\underline{C}ontrasting} (GSSC) framework to train graph data without \textbf{explicit} message passing. The GSSC framework is based purely on MLPs, where structural information is only \textbf{implicitly} incorporated to guide the computation of supervision signals. We formulate structural sparsification as a probabilistic generative model and then perform self-contrasting in the sparsified neighborhood to learn robust representations. To address the problem that the sparsification process cannot be directly optimized through back-propagation of downstream supervision, we propose a homophily-oriented objective to optimize the structural sparsification. Finally, structural sparsification and self-contrasting are formulated in a bi-level optimization framework. Despite the great progress, limitations still exist; for example exploring the applicability of the GSSC framework to other graph types, such as temporal and heterogeneous graphs, may be a promising direction for future work.

\section{Acknowledgement}
This work was supported by National Science and Technology Major Project of China (No. 2022ZD0115101), National Natural Science Foundation of China Project (No. U21A20427), and Project (No. WU2022A009) from the Center of Synthetic Biology and Integrated Bioengineering of Westlake University.

\bibliographystyle{IEEEtran}
\bibliography{Bibliography-File}

\end{document}